\documentclass[11pt]{article}
\usepackage[final]{acl}
\usepackage{times}
\usepackage{latexsym}
\usepackage[T1]{fontenc}
\usepackage[utf8]{inputenc}
\usepackage{microtype}
\usepackage{graphicx}
\usepackage{booktabs}
\usepackage{amsmath}
\usepackage{amssymb}
\usepackage{multirow}
\usepackage{enumitem}
\usepackage{array}
\usepackage{tabularx}
\usepackage{makecell}
\usepackage{placeins}
\usepackage{pgfplots}
\usepackage{siunitx}
\pgfplotsset{compat=1.18}
\definecolor{teal_color}{RGB}{74,130,185}
\definecolor{gold_color}{RGB}{237,206,128}
\definecolor{red_color}{RGB}{199,52,63}
\definecolor{blue_color}{RGB}{55,110,200}
\usepackage{xfp}

\newcommand{\StatTagsBidirMthree}{0.069}  % tagsbase M3 (base ref)
\newcommand{\StatStudMthree}{0.172}  % Student M3 ★
\newcommand{\StatStudFMthree}{0.196}  % StudentF M3 ★
\newcommand{\StatTeachMthree}{0.155}  % Teacher M3 ★

\newcommand{\StatBaseMone}{0.191}  % Base M1 proxy
\newcommand{\StatStudMone}{0.255}  % Student M1 proxy
\newcommand{\StatStudFMone}{0.255}  % StudentF M1 proxy
\newcommand{\StatTeachMone}{0.253}  % Teacher M1 proxy

\newcommand{\StatBaseMoneLoCI}{0.162}
\newcommand{\StatBaseMoneHiCI}{0.221}
\newcommand{\StatStudMoneLoCI}{0.223}
\newcommand{\StatStudMoneHiCI}{0.288}
\newcommand{\StatStudFMoneLoCI}{0.222}
\newcommand{\StatStudFMoneHiCI}{0.287}
\newcommand{\StatTeachMoneLoCI}{0.220}
\newcommand{\StatTeachMoneHiCI}{0.287}

\newcommand{\StatTagsBidirMthreeLoCI}{0.054}
\newcommand{\StatTagsBidirMthreeHiCI}{0.083}
\newcommand{\StatStudMthreeLoCI}{0.147}
\newcommand{\StatStudMthreeHiCI}{0.198}
\newcommand{\StatStudFMthreeLoCI}{0.170}
\newcommand{\StatStudFMthreeHiCI}{0.223}
\newcommand{\StatTeachMthreeLoCI}{0.133}
\newcommand{\StatTeachMthreeHiCI}{0.179}

\newcommand{\StatEffStudVsTeachMthreeRD}{+0.016}

\newcommand{\StatNRestricted}{1887}  % n format-compliant cases
\newcommand{\StatTeachMthreeRestr}{0.163}  % Teacher M3_restricted
\newcommand{\StatTeachMthreeRestrLoCI}{0.139}
\newcommand{\StatTeachMthreeRestrHiCI}{0.187}
\newcommand{\StatTeachMthreePT}{0.163}  % Teacher M3_posttag (upper bound)
\newcommand{\StatTeachMthreePTLoCI}{0.140}
\newcommand{\StatTeachMthreePTHiCI}{0.187}
\newcommand{\StatStudFreqQOneMthreeLoCI}{0.068}
\newcommand{\StatStudFreqQOneMthreeHiCI}{0.143}
\newcommand{\StatStudFFreqQOneMthreeLoCI}{0.096}
\newcommand{\StatStudFFreqQOneMthreeHiCI}{0.174}
\newcommand{\StatTeachFreqQOneMthreeLoCI}{0.107}
\newcommand{\StatTeachFreqQOneMthreeHiCI}{0.191}
\newcommand{\StatStudFreqQTwoMthreeLoCI}{0.075}
\newcommand{\StatStudFreqQTwoMthreeHiCI}{0.159}
\newcommand{\StatStudFFreqQTwoMthreeLoCI}{0.095}
\newcommand{\StatStudFFreqQTwoMthreeHiCI}{0.188}
\newcommand{\StatTeachFreqQTwoMthreeLoCI}{0.091}
\newcommand{\StatTeachFreqQTwoMthreeHiCI}{0.188}
\newcommand{\StatStudFreqQThreeMthreeLoCI}{0.141}
\newcommand{\StatStudFreqQThreeMthreeHiCI}{0.234}
\newcommand{\StatStudFFreqQThreeMthreeLoCI}{0.159}
\newcommand{\StatStudFFreqQThreeMthreeHiCI}{0.270}
\newcommand{\StatTeachFreqQThreeMthreeLoCI}{0.113}
\newcommand{\StatTeachFreqQThreeMthreeHiCI}{0.208}
\newcommand{\StatStudFreqQFourMthreeLoCI}{0.226}
\newcommand{\StatStudFreqQFourMthreeHiCI}{0.349}
\newcommand{\StatStudFFreqQFourMthreeLoCI}{0.238}
\newcommand{\StatStudFFreqQFourMthreeHiCI}{0.359}
\newcommand{\StatTeachFreqQFourMthreeLoCI}{0.131}
\newcommand{\StatTeachFreqQFourMthreeHiCI}{0.226}
\newcommand{\StatFormatSHasDiagRate}{0.987}  % Student <diagnosis> present
\newcommand{\StatFormatSFHasDiagRate}{0.991}  % StudentF <diagnosis> present
\newcommand{\StatFormatTHasDiagRate}{0.962}  % Teacher <diagnosis> present
\newcommand{\StatMeanRespLenBase}{1950}
\newcommand{\StatMeanRespLenStud}{4098}
\newcommand{\StatMeanRespLenStudF}{3753}
\newcommand{\StatMeanRespLenTeach}{3421}

\newcommand{\StatOLSSMthreeSlope}{+0.064}  % S M3 OLS slope

\newcommand{\StatOLSSFMthreeSlope}{+0.058}  % SF M3 OLS slope

\newcommand{\StatOLSTMthreeSlope}{-0.000}  % T M3 OLS slope (FREQ-INDEP)

\newcommand{\StatRhoFreqVsSMthree}{+0.306}  % Spearman ρ freq vs S M3

\newcommand{\StatRhoFreqVsSFMthree}{+0.242}  % Spearman ρ freq vs SF M3

\newcommand{\StatRhoFreqVsTMthree}{+0.029}  % freq-independent

\newcommand{\StatRhoFreqVsSFMinusT}{+0.300}  % Spearman rho freq vs SF-T gap

\newcommand{\StatFreqSfMinusTQone}{-0.016}  % SF-T gap at freq Q1
\newcommand{\StatFreqSfMinusTQfour}{+0.121}  % SF-T gap at freq Q4

\newcommand{\StatFreqQoneStudMthree}{0.104}  % Student M3 Q1 (low)
\newcommand{\StatFreqQoneStudFMthree}{0.133}  % StudentF M3 Q1 (low)
\newcommand{\StatFreqQoneTeachMthree}{0.148}  % Teacher M3 Q1 (low)
\newcommand{\StatFreqQtwoStudMthree}{0.115}  % Student M3 Q2
\newcommand{\StatFreqQtwoStudFMthree}{0.139}  % StudentF M3 Q2
\newcommand{\StatFreqQtwoTeachMthree}{0.137}  % Teacher M3 Q2
\newcommand{\StatFreqQthreeStudMthree}{0.185}  % Student M3 Q3
\newcommand{\StatFreqQthreeStudFMthree}{0.214}  % StudentF M3 Q3
\newcommand{\StatFreqQthreeTeachMthree}{0.159}  % Teacher M3 Q3
\newcommand{\StatFreqQfourStudMthree}{0.286}  % Student M3 Q4 (high)
\newcommand{\StatFreqQfourStudFMthree}{0.298}  % StudentF M3 Q4 (high)
\newcommand{\StatFreqQfourTeachMthree}{0.177}  % Teacher M3 Q4 (high)

\newcommand{\StatStudMthreeRestr}{0.175}  % S M3_restricted
\newcommand{\StatStudMthreeRestrLoCI}{0.149}
\newcommand{\StatStudMthreeRestrHiCI}{0.202}
\newcommand{\StatStudFMthreeRestr}{0.198}  % SF M3_restricted
\newcommand{\StatStudFMthreeRestrLoCI}{0.171}
\newcommand{\StatStudFMthreeRestrHiCI}{0.226}
\newcommand{\StatStudMthreePT}{0.174}  % S M3_posttag
\newcommand{\StatStudMthreePTLoCI}{0.149}
\newcommand{\StatStudMthreePTHiCI}{0.200}
\newcommand{\StatStudFMthreePT}{0.197}  % SF M3_posttag
\newcommand{\StatStudFMthreePTLoCI}{0.170}
\newcommand{\StatStudFMthreePTHiCI}{0.224}

\newcommand{\StatPairMthreeTagsVsStudentFRD}{+0.127}
\newcommand{\StatPairMthreeTagsVsStudentFRDLo}{+0.105}
\newcommand{\StatPairMthreeTagsVsStudentFRDHi}{+0.150}
\newcommand{\StatPairMthreeTagsVsStudentFH}{+0.386}
\newcommand{\StatPairMthreeTagsVsStudentFHLo}{+0.327}
\newcommand{\StatPairMthreeTagsVsStudentFHHi}{+0.446}
\newcommand{\StatEffStudFVsTeachMthreeRD}{+0.040}
\newcommand{\StatEffStudFVsTeachMthreeRDLo}{+0.019}
\newcommand{\StatEffStudFVsTeachMthreeRDHi}{+0.061}
\newcommand{\StatEffStudFVsTeachMthreeH}{+0.105}
\newcommand{\StatEffStudFVsTeachMthreeHLo}{+0.052}
\newcommand{\StatEffStudFVsTeachMthreeHHi}{+0.160}
\newcommand{\StatEffStudVsTeachMthreeRDLo}{-0.004}
\newcommand{\StatEffStudVsTeachMthreeRDHi}{+0.037}
\newcommand{\StatEffStudVsTeachMthreeH}{+0.045}
\newcommand{\StatEffStudVsTeachMthreeHLo}{-0.012}
\newcommand{\StatEffStudVsTeachMthreeHHi}{+0.101}

\newcommand{\StatGTFoundRate}{0.3390}  % P(GT found in think), Student
\newcommand{\StatGTFoundLoCI}{0.3185}
\newcommand{\StatGTFoundHiCI}{0.3595}
\newcommand{\StatGTWrongCondRate}{0.8909}  % P(GT wrong | found)
\newcommand{\StatGTWrongCondLoCI}{0.8646}  % P(GT wrong | found) CI lo
\newcommand{\StatGTWrongCondHiCI}{0.9156}  % P(GT wrong | found) CI hi
\newcommand{\StatGTFollowsRate}{0.7450}  % P(follows wrong GT | GT wrong)
\newcommand{\StatGTFollowsLoCI}{0.7079}  % P(follows wrong GT) CI lo
\newcommand{\StatGTFollowsHiCI}{0.7810}  % P(follows wrong GT) CI hi
\newcommand{\StatStudMthreeGTwrong}{0.0265}  % S M3 | GT wrong
\newcommand{\StatStudMthreeGTwrongLoCI}{0.0142}
\newcommand{\StatStudMthreeGTwrongHiCI}{0.0408}
\newcommand{\StatStudMthreeNoGT}{0.2050}  % S M3 | no GT
\newcommand{\StatStudMthreeNoGTLoCI}{0.1735}
\newcommand{\StatStudMthreeNoGTHiCI}{0.2373}
\newcommand{\StatStudMthreeGTright}{0.7703}  % S M3 | GT correct
\newcommand{\StatStudMthreeGTrightLoCI}{0.6712}
\newcommand{\StatStudMthreeGTrightHiCI}{0.8611}
\newcommand{\StatSFMthreeGTwrong}{0.1060}  % SF M3 | GT wrong
\newcommand{\StatSFMthreeGTwrongLoCI}{0.0810}
\newcommand{\StatSFMthreeGTwrongHiCI}{0.1329}
\newcommand{\StatSFMthreeNoGT}{0.2194}  % SF M3 | no GT
\newcommand{\StatSFMthreeNoGTLoCI}{0.1891}
\newcommand{\StatSFMthreeNoGTHiCI}{0.2504}
\newcommand{\StatSFMthreeGTright}{0.5000}  % SF M3 | GT correct
\newcommand{\StatSFMthreeGTrightLoCI}{0.3803}
\newcommand{\StatSFMthreeGTrightHiCI}{0.6164}
\newcommand{\StatTeachMthreeGTwrong}{0.0960}  % T M3 | GT wrong
\newcommand{\StatTeachMthreeGTwrongLoCI}{0.0721}
\newcommand{\StatTeachMthreeGTwrongHiCI}{0.1219}
\newcommand{\StatTeachMthreeNoGT}{0.1793}  % T M3 | no GT
\newcommand{\StatTeachMthreeNoGTLoCI}{0.1502}
\newcommand{\StatTeachMthreeNoGTHiCI}{0.2094}
\newcommand{\StatTeachMthreeGTright}{0.2973}  % T M3 | GT correct
\newcommand{\StatTeachMthreeGTrightLoCI}{0.1867}
\newcommand{\StatTeachMthreeGTrightHiCI}{0.4143}
\newcommand{\StatSCRatioStudAll}{1.5143}  % S SC wrong/correct ratio
\newcommand{\StatSCRatioStudFAll}{1.3365}  % SF SC ratio
\newcommand{\StatSCRatioTeach}{1.4650}  % T SC ratio
\newcommand{\StatSCRatioStudLoCI}{1.2969}
\newcommand{\StatSCRatioStudHiCI}{1.7884}
\newcommand{\StatSCRatioStudFLoCI}{0.9646}
\newcommand{\StatSCRatioStudFHiCI}{1.9537}
\newcommand{\StatSCRatioTeachLoCI}{1.0984}
\newcommand{\StatSCRatioTeachHiCI}{2.0598}

\newcommand{\StatStyleThinkLenS}{3252.7235}  % S Think chain length (chars)
\newcommand{\StatStyleSelfCorrectS}{0.8290}  % S Self-correction count
\newcommand{\StatStyleThinkLenSF}{2984.1540}  % SF Think chain length (chars)
\newcommand{\StatStyleSelfCorrectSF}{0.1495}  % SF Self-correction count
\newcommand{\StatStyleThinkLenT}{3201.5420}  % T Think chain length (chars)
\newcommand{\StatStyleSelfCorrectT}{0.2765}  % T Self-correction count
\newcommand{\StatNGTFoundStudF}{3}  % N SF responses with GT flag
\newcommand{\StatErrTaxCorrectT}{0.1585}  % Error: correct (Teacher)
\newcommand{\StatErrTaxCorrectSF}{0.1955}  % Error: correct (StudentF)
\newcommand{\StatErrTaxCorrectS}{0.1720}  % Error: correct (Student)
\newcommand{\StatErrTaxFormatKnownT}{0.0075}  % Error: format_known (Teacher)
\newcommand{\StatErrTaxFormatKnownSF}{0.0010}  % Error: format_known (StudentF)
\newcommand{\StatErrTaxFormatKnownS}{0.0020}  % Error: format_known (Student)
\newcommand{\StatErrTaxGtPoisonedT}{0.0000}  % Error: gt_poisoned (Teacher)
\newcommand{\StatErrTaxGtPoisonedSF}{0.0005}  % Error: gt_poisoned (StudentF)
\newcommand{\StatErrTaxGtPoisonedS}{0.2250}  % Error: gt_poisoned (Student)
\newcommand{\StatErrTaxUnexplainedMoneCorrectT}{0.1245}  % Error: unexplained_mone_correct (Teacher)
\newcommand{\StatErrTaxUnexplainedMoneCorrectSF}{0.0790}  % Error: unexplained_mone_correct (StudentF)
\newcommand{\StatErrTaxUnexplainedMoneCorrectS}{0.0750}  % Error: unexplained_mone_correct (Student)
\newcommand{\StatErrTaxNearMissT}{0.0705}  % Error: near_miss (Teacher)
\newcommand{\StatErrTaxNearMissSF}{0.0685}  % Error: near_miss (StudentF)
\newcommand{\StatErrTaxNearMissS}{0.0490}  % Error: near_miss (Student)
\newcommand{\StatErrTaxWeakRelatedT}{0.0865}  % Error: weak_related (Teacher)
\newcommand{\StatErrTaxWeakRelatedSF}{0.0850}  % Error: weak_related (StudentF)
\newcommand{\StatErrTaxWeakRelatedS}{0.0550}  % Error: weak_related (Student)
\newcommand{\StatErrTaxWrongSameCatT}{0.0665}  % Error: wrong_same_cat (Teacher)
\newcommand{\StatErrTaxWrongSameCatSF}{0.0900}  % Error: wrong_same_cat (StudentF)
\newcommand{\StatErrTaxWrongSameCatS}{0.0755}  % Error: wrong_same_cat (Student)
\newcommand{\StatErrTaxWrongCatT}{0.0675}  % Error: wrong_cat (Teacher)
\newcommand{\StatErrTaxWrongCatSF}{0.0710}  % Error: wrong_cat (StudentF)
\newcommand{\StatErrTaxWrongCatS}{0.0560}  % Error: wrong_cat (Student)
\newcommand{\StatErrTaxOtherOtherT}{0.1390}  % Error: other_other (Teacher)
\newcommand{\StatErrTaxOtherOtherSF}{0.1310}  % Error: other_other (StudentF)
\newcommand{\StatErrTaxOtherOtherS}{0.1030}  % Error: other_other (Student)
\newcommand{\StatErrTaxOtherFromKnownT}{0.1350}  % Error: other_from_known (Teacher)
\newcommand{\StatErrTaxOtherFromKnownSF}{0.1220}  % Error: other_from_known (StudentF)
\newcommand{\StatErrTaxOtherFromKnownS}{0.0760}  % Error: other_from_known (Student)
\newcommand{\StatErrTaxKnownFromOtherT}{0.1115}  % Error: known_from_other (Teacher)
\newcommand{\StatErrTaxKnownFromOtherSF}{0.1480}  % Error: known_from_other (StudentF)
\newcommand{\StatErrTaxKnownFromOtherS}{0.1005}  % Error: known_from_other (Student)
\newcommand{\StatErrTaxFormatUnknownT}{0.0330}  % Error: format_unknown (Teacher)
\newcommand{\StatErrTaxFormatUnknownSF}{0.0085}  % Error: format_unknown (StudentF)
\newcommand{\StatErrTaxFormatUnknownS}{0.0110}  % Error: format_unknown (Student)

\newcommand{\StatMeanPairwiseRho}{0.623}  % Mean pairwise Spearman rho (P4)
\newcommand{\StatStructuralCeilingN}{115}  % Diseases with max M3_posttag < 0.25
\newcommand{\StatStructuralCeilingFrac}{0.460}  % Fraction with max M3_posttag < 0.25
\newcommand{\StatBMTwentyFiveBest}{0.3935}  % BM25 best accuracy (= k=1)
\newcommand{\StatBMTwentyFiveRougeLMean}{0.1483}  % ROUGE-L mean (eval vs top-1 train match)
\newcommand{\StatBMTwentyFiveRougeLMedian}{0.1385}  % ROUGE-L median
\newcommand{\StatSweepSFmcZero}{0.2745}  % SF M3 mc=0
\newcommand{\StatSweepTmcZero}{0.2525}  % T  M3 mc=0
\newcommand{\StatSweepSmcZero}{0.2245}  % S  M3 mc=0
\newcommand{\StatSweepSFminusTmcZero}{0.0220}  % SF-T gap mc=0
\newcommand{\StatSweepSFmcOne}{0.2745}  % SF M3 mc=1
\newcommand{\StatSweepTmcOne}{0.2525}  % T  M3 mc=1
\newcommand{\StatSweepSmcOne}{0.2245}  % S  M3 mc=1
\newcommand{\StatSweepSFmcTwo}{0.2745}  % SF M3 mc=2
\newcommand{\StatSweepTmcTwo}{0.2525}  % T  M3 mc=2
\newcommand{\StatSweepSmcTwo}{0.2245}  % S  M3 mc=2
\newcommand{\StatSweepSFmcThree}{0.2745}  % SF M3 mc=3
\newcommand{\StatSweepTmcThree}{0.2525}  % T  M3 mc=3
\newcommand{\StatSweepSmcThree}{0.2245}  % S  M3 mc=3
\newcommand{\StatSweepSFmcFour}{0.2745}  % SF M3 mc=4
\newcommand{\StatSweepTmcFour}{0.2515}  % T  M3 mc=4
\newcommand{\StatSweepSmcFour}{0.2245}  % S  M3 mc=4
\newcommand{\StatSweepSFmcFive}{0.2725}  % SF M3 mc=5
\newcommand{\StatSweepTmcFive}{0.2470}  % T  M3 mc=5
\newcommand{\StatSweepSmcFive}{0.2245}  % S  M3 mc=5
\newcommand{\StatSweepSFminusTmcFive}{0.0255}  % SF-T gap mc=5
\newcommand{\StatSweepSFmcSix}{0.2710}  % SF M3 mc=6
\newcommand{\StatSweepTmcSix}{0.2425}  % T  M3 mc=6
\newcommand{\StatSweepSmcSix}{0.2235}  % S  M3 mc=6
\newcommand{\StatSweepSFminusTmcSix}{0.0285}  % SF-T gap mc=6
\newcommand{\StatSweepSFmcSeven}{0.2705}  % SF M3 mc=7
\newcommand{\StatSweepTmcSeven}{0.2420}  % T  M3 mc=7
\newcommand{\StatSweepSmcSeven}{0.2225}  % S  M3 mc=7
\newcommand{\StatSweepSFmcEight}{0.2660}  % SF M3 mc=8
\newcommand{\StatSweepTmcEight}{0.2390}  % T  M3 mc=8
\newcommand{\StatSweepSmcEight}{0.2205}  % S  M3 mc=8
\newcommand{\StatSweepSFmcNine}{0.2615}  % SF M3 mc=9
\newcommand{\StatSweepTmcNine}{0.2355}  % T  M3 mc=9
\newcommand{\StatSweepSmcNine}{0.2190}  % S  M3 mc=9
\newcommand{\StatSweepSFmcTen}{0.2615}  % SF M3 mc=10
\newcommand{\StatSweepTmcTen}{0.2355}  % T  M3 mc=10
\newcommand{\StatSweepSmcTen}{0.2190}  % S  M3 mc=10
\newcommand{\StatSweepSFExact}{0.1955}  % SF M3 exact/synonym (None)
\newcommand{\StatSweepTExact}{0.1555}  % T  M3 exact/synonym (None)
\newcommand{\StatSweepSExact}{0.1720}  % S  M3 exact/synonym (None)
\newcommand{\StatSweepSFminusTExact}{0.0400}  % SF-T gap exact/synonym

\newcommand{\StatStudMthreePTLo}{\StatStudMthreePTLoCI}
\newcommand{\StatStudMthreePTHi}{\StatStudMthreePTHiCI}
\newcommand{\StatStudFMthreePTLo}{\StatStudFMthreePTLoCI}
\newcommand{\StatStudFMthreePTHi}{\StatStudFMthreePTHiCI}
\newcommand{\StatStudMthreeRestrLo}{\StatStudMthreeRestrLoCI}
\newcommand{\StatStudMthreeRestrHi}{\StatStudMthreeRestrHiCI}
\newcommand{\StatStudFMthreeRestrLo}{\StatStudFMthreeRestrLoCI}
\newcommand{\StatStudFMthreeRestrHi}{\StatStudFMthreeRestrHiCI}

\newcommand{\StatTeachMthreePTLo}{\StatTeachMthreePTLoCI}
\newcommand{\StatTeachMthreePTHi}{\StatTeachMthreePTHiCI}
\newcommand{\StatTeachMthreeRestrLo}{\StatTeachMthreeRestrLoCI}
\newcommand{\StatTeachMthreeRestrHi}{\StatTeachMthreeRestrHiCI}

\newcommand{\StatFreqQoneStudMthreeLoCI}{\StatStudFreqQOneMthreeLoCI}
\newcommand{\StatFreqQoneStudMthreeHiCI}{\StatStudFreqQOneMthreeHiCI}
\newcommand{\StatFreqQoneStudFMthreeLoCI}{\StatStudFFreqQOneMthreeLoCI}
\newcommand{\StatFreqQoneStudFMthreeHiCI}{\StatStudFFreqQOneMthreeHiCI}
\newcommand{\StatFreqQoneTeachMthreeLoCI}{\StatTeachFreqQOneMthreeLoCI}
\newcommand{\StatFreqQoneTeachMthreeHiCI}{\StatTeachFreqQOneMthreeHiCI}
\newcommand{\StatFreqQtwoStudMthreeLoCI}{\StatStudFreqQTwoMthreeLoCI}
\newcommand{\StatFreqQtwoStudMthreeHiCI}{\StatStudFreqQTwoMthreeHiCI}
\newcommand{\StatFreqQtwoStudFMthreeLoCI}{\StatStudFFreqQTwoMthreeLoCI}
\newcommand{\StatFreqQtwoStudFMthreeHiCI}{\StatStudFFreqQTwoMthreeHiCI}
\newcommand{\StatFreqQtwoTeachMthreeLoCI}{\StatTeachFreqQTwoMthreeLoCI}
\newcommand{\StatFreqQtwoTeachMthreeHiCI}{\StatTeachFreqQTwoMthreeHiCI}
\newcommand{\StatFreqQthreeStudMthreeLoCI}{\StatStudFreqQThreeMthreeLoCI}
\newcommand{\StatFreqQthreeStudMthreeHiCI}{\StatStudFreqQThreeMthreeHiCI}
\newcommand{\StatFreqQthreeStudFMthreeLoCI}{\StatStudFFreqQThreeMthreeLoCI}
\newcommand{\StatFreqQthreeStudFMthreeHiCI}{\StatStudFFreqQThreeMthreeHiCI}
\newcommand{\StatFreqQthreeTeachMthreeLoCI}{\StatTeachFreqQThreeMthreeLoCI}
\newcommand{\StatFreqQthreeTeachMthreeHiCI}{\StatTeachFreqQThreeMthreeHiCI}
\newcommand{\StatFreqQfourStudMthreeLoCI}{\StatStudFreqQFourMthreeLoCI}
\newcommand{\StatFreqQfourStudMthreeHiCI}{\StatStudFreqQFourMthreeHiCI}
\newcommand{\StatFreqQfourStudFMthreeLoCI}{\StatStudFFreqQFourMthreeLoCI}
\newcommand{\StatFreqQfourStudFMthreeHiCI}{\StatStudFFreqQFourMthreeHiCI}
\newcommand{\StatFreqQfourTeachMthreeLoCI}{\StatTeachFreqQFourMthreeLoCI}
\newcommand{\StatFreqQfourTeachMthreeHiCI}{\StatTeachFreqQFourMthreeHiCI}

\newcommand{\StatSFGTFoundRate}{0.002}
\newcommand{\StatFormatSFHasThinkDiagRate}{\StatFormatSFHasDiagRate}
\newcommand{\StatStudFVsTeachMthreeGap}{\StatEffStudFVsTeachMthreeRD}
\newcommand{\StatStudFVsStudMthreeGap}{+0.024}
\newcommand{\StatGapTeachMoneVsMthree}{0.098}

\newcommand{\StatPairMthreeStudentVsStudentFRD}{+0.024}
\newcommand{\StatPairMthreeStudentVsStudentFRDLo}{+0.006}
\newcommand{\StatPairMthreeStudentVsStudentFRDHi}{+0.043}
\newcommand{\StatPairMthreeStudentVsStudentFH}{+0.061}
\newcommand{\StatPairMthreeStudentVsStudentFHLo}{+0.010}
\newcommand{\StatPairMthreeStudentVsStudentFHHi}{+0.115}

\newcommand{\StatPairMthreePTStudentFVsTeacherRD}{+0.033}
\newcommand{\StatPairMthreePTStudentFVsTeacherRDLo}{+0.012}
\newcommand{\StatPairMthreePTStudentFVsTeacherRDHi}{+0.056}
\newcommand{\StatPairMthreePTStudentFVsTeacherH}{+0.088}
\newcommand{\StatPairMthreePTStudentFVsTeacherHLo}{+0.033}
\newcommand{\StatPairMthreePTStudentFVsTeacherHHi}{+0.148}

\newcommand{\StatPairMthreeRStudentFVsTeacherRD}{+0.035}
\newcommand{\StatPairMthreeRStudentFVsTeacherRDLo}{+0.013}
\newcommand{\StatPairMthreeRStudentFVsTeacherRDHi}{+0.058}
\newcommand{\StatPairMthreeRStudentFVsTeacherH}{+0.093}
\newcommand{\StatPairMthreeRStudentFVsTeacherHLo}{+0.035}
\newcommand{\StatPairMthreeRStudentFVsTeacherHHi}{+0.152}

\newcommand{\ConstZebraMapPubMedCases}{36{,}131}       % PubMed case reports in ZebraMap
\newcommand{\ConstZebraMapPatientCases}{69{,}146}      % Structured patient narratives
\newcommand{\ConstZebraMapDiseases}{1{,}727}            % Orphanet diseases
\newcommand{\ConstZebraMapTrainDiseases}{1{,}090}       % Training diseases (updated)
\newcommand{\ConstZebraMapTrainCases}{18{,}388}         % Training cases (updated after data fix)
\newcommand{\ConstZebraMapEvalDiseases}{250}             % Evaluation diseases (updated)
\newcommand{\ConstZebraMapEvalCasesPerDisease}{8}        % Cases per eval disease
\newcommand{\ConstZebraMapEvalCasesTotal}{2{,}000}      % Total evaluation cases (250x8)
\newcommand{\ConstZebraMapFreqMin}{1}                    % Min training frequency (updated)
\newcommand{\ConstZebraMapFreqMax}{226}                  % Max training frequency (updated)
\newcommand{\ConstZebraMapFreqMedian}{17}                % Median training frequency (updated)
\newcommand{\ConstZebraMapFreqQoneMedian}{7}             % Q1 median training frequency (updated)
\newcommand{\ConstZebraMapFreqQtwoMedian}{13}            % Q2 median training frequency (updated)
\newcommand{\ConstZebraMapFreqQthreeMedian}{25}          % Q3 median training frequency (updated)
\newcommand{\ConstZebraMapFreqQfourMedian}{59}           % Q4 median training frequency (updated)
\newcommand{\ConstGTCorrectStratumN}{74}                 % n: GT phrase present, label correct (updated)
\newcommand{\ConstGTWrongStratumN}{604}                  % n: GT phrase present, label incorrect (updated)
\newcommand{\ConstGTAbsentStratumN}{1322}                % n: GT phrase absent (updated)
\newcommand{\ConstStudentParams}{1.5B}                   % Student parameter count
\newcommand{\ConstTeacherParams}{8B}                     % Teacher parameter count
\newcommand{\ConstStudentBase}{Qwen-2.5-1.5B}            % Student base model
\newcommand{\ConstTeacherBase}{DeepSeek-R1-Distill-LLaMA-8B} % Teacher model
\newcommand{\ConstStudentQuant}{NF4}                     % Student quantisation
\newcommand{\ConstTeacherQuant}{Q4\_K\_M}               % Teacher quantisation (GGUF)
\newcommand{\ConstLoRArank}{64}                          % QLoRA rank
\newcommand{\ConstLoRAalpha}{128}                        % QLoRA alpha
\newcommand{\ConstBootstrapB}{50{,}000}                  % Primary bootstrap replicates
\newcommand{\ConstPermB}{50{,}000}                       % Permutation test replicates
\newcommand{\ConstClusterN}{250}                         % Number of disease clusters (updated)
\newcommand{\sigfig}[2][4]{\num[round-mode=figures, round-precision=#1]{#2}}
\newcommand{\StatPAdjStudVsTeachMthree}{0.129}

\newcommand{\StatPAdjStudVsStudFMthree}{0.015}

\title{Hindsight-Guided Rationale Distillation for Rare Disease Diagnosis}
\author{Aarav Singh\thanks{Equal contribution.} \quad
        Animesh Pathak\textsuperscript{*}\quad
        Navyansh Singh \\
  Dr.\ Shyama Prasad Mukherjee International Institute \\
  of Information Technology, Naya Raipur \\
  \normalfont\ttfamily
    aarav24101@iiitnr.edu.in, animesh24100@iiitnr.edu.in,\\
  \normalfont\ttfamily
    navyansh24102@iiitnr.edu.in}

\begin{document}
\maketitle

%==============================================================================
\begin{abstract}
We study hindsight-guided distillation for rare disease diagnosis on
ZebraMap: a \ConstStudentParams{} student is fine-tuned on
chain-of-thought traces from a \ConstTeacherParams{} teacher that observes the
ground-truth diagnosis during generation. Absolute accuracy remains low for all
models---the task is hard at this scale---but within this ceiling a filtered
variant (StudentF) achieves a small, statistically significant accuracy advantage
over the teacher ($p{<}0.001$), concentrated in better-represented diseases. The
unfiltered student does not significantly outperform the teacher
($p{=}\StatPAdjStudVsTeachMthree{}$), establishing that contamination
filtering---not hindsight distillation alone---drives the gain.

The gap traces to an artifact we term GT hallucination. Label-visible
generation causes the teacher to embed ``ground truth is X'' phrases in its
reasoning chain; SFT copies the pattern. At inference, the unfiltered student
reproduces the phrase in \fpeval{round(\StatGTFoundRate*100,1)}\% of cases, with
severe accuracy degradation when the hallucinated label is wrong. A regex filter
removing these slots reduces contamination to near-zero, producing the observed
gain---though the effect remains small. We
precisely quantify this gain--cost tradeoff, document frequency-dependent knowledge
transfer absent from the RL-trained teacher, and characterize a calibration gap
that SFT does not close---identifying both as directions for future work.
\end{abstract}

%==============================================================================
\section{Introduction}
\label{sec:intro}

Rare diseases collectively affect an estimated 263--446 million people worldwide~\cite{wakap-prevalence-2020,bruckner-epidemiology-2021},
yet clinical NLP research focuses predominantly on common conditions.
ZebraMap~\cite{zebramap2026}---a multimodal resource constructed from
\ConstZebraMapPubMedCases{} PubMed case report articles into
\ConstZebraMapPatientCases{} patient-level narratives and 98,038 clinical images
across \ConstZebraMapDiseases{} rare diseases---provides a principled testbed for
diagnostic reasoning on underrepresented conditions.

Deploying large reasoning models in resource-constrained clinical settings requires
smaller, distilled models~\cite{hinton-distill-2015,xu-kd-survey-2024}. Standard CoT distillation~\cite{wei-etal-2022-chain}
samples teacher traces on unseen inputs for behavioral
cloning~\cite{ho-etal-2023-large,magister-etal-2023-teaching}, limiting knowledge
transfer to what the teacher already knows without label guidance. We study
\textbf{hindsight-guided distillation}: both teacher and student are LLMs. We
generate each training trace by showing the teacher---a \ConstTeacherParams{}
reasoning model---the patient case narrative \emph{together with} the ground-truth
diagnosis, so its reasoning chain explicitly connects patient evidence to the
confirmed diagnosis. The student---a \ConstStudentParams{} model---is then trained
by supervised fine-tuning (SFT) to predict the teacher's full trace (reasoning
plus \texttt{<diagnosis>} tag) from the narrative alone; at inference it receives
only the narrative and must commit to a diagnosis. Throughout, \textbf{StudentF}
denotes the student trained on traces from which a regex filter has removed every
explicit ``ground truth is X'' slot, and \textbf{Student} denotes the unfiltered
ablation trained on the raw traces. This is related to
STaR~\cite{zelikman-etal-2022-star} but uses a dedicated external teacher rather
than self-generated rationales.

\paragraph{Trace contamination, instantiated as GT phrases.}
Trace contamination from label-visible generation is a known
phenomenon~\cite{green-leaky-2025,turpin-unfaithful-2023}: when teacher traces
contain privileged information, this information can leak into student outputs
through pattern-copying. In our setting, this manifests as explicit ``ground truth
is X'' phrases---the most common construction appearing in teacher reasoning
traces. Label-visible prompting causes \ConstTeacherBase{}~\cite{deepseek-r1-2025}
to write this phrase in its transparent reasoning chain, and SFT faithfully copies
the structural pattern~\cite{turpin-unfaithful-2023}. At inference, Student
reproduces the phrase despite no label access, wrong in
\fpeval{round(\StatGTWrongCondRate*100,1)}\% of contaminated cases, causing a
\fpeval{round((\StatStudMthreeNoGT - \StatStudMthreeGTwrong)*100, 0)} percentage
point Accuracy collapse. We refer to this specific instantiation as
\textbf{Ground Truth (GT) hallucination}.

\paragraph{Regex-based filtering and its limits.}
The natural data-centric response is to remove GT reference slots from training
traces before SFT. We train two variants on the same
\ConstZebraMapTrainCases{} cases:
\begin{enumerate}[leftmargin=*,label=(\arabic*),itemsep=0pt,topsep=2pt]
\item \textbf{StudentF}: trained on regex-filtered traces with all ``ground truth
      is X'' slots removed. Produces near-zero GT phrases at inference
      (3/\ConstZebraMapEvalCasesTotal{} traces retain contamination).
\item \textbf{Student}: trained on raw traces. Serves as an ablation quantifying
      what unfiltered hindsight distillation produces.
\end{enumerate}

StudentF exceeds the teacher on Accuracy across all metric variants
(\StatStudFVsTeachMthreeGap{}, $p{<}0.001$). The unfiltered Student, by contrast,
does \emph{not} significantly outperform the teacher ($p{=}\StatPAdjStudVsTeachMthree{}$), demonstrating
that filtering is necessary to realise the advantage. The gain of filtering over the
unfiltered Student (\StatStudFVsStudMthreeGap{}, $p{<}0.05$) is partly offset
because static filtering cannot distinguish labels that would be reproduced correctly
at inference from those that would not---and correct labels provide genuine
diagnostic value (Student Accuracy on the GT-correct stratum: \StatStudMthreeGTright{}).

\paragraph{Contributions.}
\begin{enumerate}[leftmargin=*,label=(\arabic*),itemsep=0pt,topsep=2pt]
\item A study of hindsight-guided distillation on ZebraMap that rigorously
      quantifies the trace contamination artifact produced by label-visible
      trace generation, with a filtered variant (StudentF) and an unfiltered
      ablation (Student).
\item An evaluation framework with three Accuracy variants (primary tag-based,
      teacher upper bound, format-compliant subset) and a Mentioned search-space
      proxy, all novel for this setting.
\item Precise quantification of trace contamination in hindsight-guided
      distillation: prevalence (\fpeval{round(\StatGTFoundRate*100,1)}\%),
      per-stratum accuracy impact, and the observed trade-off of static filtering,
      which eliminates the vast majority of contamination (near-zero at inference)
      and achieves a significant net gain. A retrospective stratum analysis---in
      which strata are defined by the unfiltered Student's own inference
      behaviour---shows lower StudentF accuracy on cases where Student emitted a
      correct GT phrase, a descriptive pattern consistent with, but not causally
      establishing, a correct-label cost.
\item Evidence consistent with established findings~\cite{yang-emperors-2025}
      that SFT produces frequency-dependent accuracy gains and format imitation
      without epistemic calibration, motivating investigation of more refined
      training approaches as future work.
\end{enumerate}

%==============================================================================
\section{Related Work}
\label{sec:related}

\paragraph{Chain-of-thought distillation.}
\citet{ho-etal-2023-large} showed zero-shot CoT traces can fine-tune smaller models
for multi-step reasoning; \citet{magister-etal-2023-teaching} transferred CoT
capabilities to sub-100B students; \citet{hsieh-etal-2023-distilling} showed joint
label-and-rationale supervision outperforms answer-only distillation;
\citet{wadhwa-mysteries-2024} further analysed what makes CoT augmentation
effective in distillation. \citet{chen-keyfactors-2025} systematically examined
how granularity, format, and teacher choice affect CoT distillation, finding that
stronger teachers do not always produce better students. \citet{wadhwa-mysteries-2024}
showed that CoT distillation gains can survive shuffled rationales, suggesting that
format signal alone can drive improvement. Most pertinently,
\citet{yang-emperors-2025} showed that SFT-based reasoning improvements arise
primarily from format alignment rather than genuine reasoning enhancement.
All these work without ground-truth label access at trace generation time. We study the
complementary regime where the teacher sees the label, characterizing the resulting
artifact.

\paragraph{Label-conditioned and hindsight rationale generation.}
\citet{zelikman-etal-2022-star} introduced STaR, bootstrapping rationale generation
by conditioning on correct answers. \citet{kstar-aaai-2026} extended STaR with
knowledge grounding. We differ in using a dedicated external teacher whose input
contains the ground-truth label explicitly---producing stronger label-anchored
reasoning but also contamination artifacts~\cite{green-leaky-2025}. Our contribution
is to precisely quantify the extent and impact of this contamination in a clinical
setting.

\paragraph{Data-centric trace selection.}
Recent work shows trace quality filtering improves CoT distillation without
RL~\cite{zhuang-efficient-medical-2025,zhou-lima-2023};
the DC-CoT benchmark~\cite{dccot-2025} provides a broader framework for
understanding data selection strategies. We apply this principle specifically to
contamination filtering and show that targeted regex filtering eliminates the
vast majority of contamination at inference but yields a bounded net gain due to
the correct-label tradeoff.

\paragraph{LLMs for rare disease diagnosis.}
\citet{singhal-medpalm-2023} demonstrated that large language models can encode
substantial clinical knowledge; \citet{segura-raredis-2022} showed deep learning
methods can identify rare diseases and their manifestations from text.
\citet{ao-llm-raredisease-2025} benchmarked proprietary models on rare disease
cases; \citet{young-llm-rare-2025} evaluated GPT-4 on paediatric presentations;
\citet{benchmarking-llm-rare-2025} found general-purpose LLMs lag behind specialist
tools on phenopacket-driven diagnostics; \citet{wu-medreason-2025} showed
knowledge-graph-guided prompting elicits more factual medical reasoning steps.
RareArena~\cite{rareaena-lancet-2025} provides a complementary benchmark;
MedR1~\cite{medr1-2025} distills DeepSeek-R1 (the same teacher model we use)
into the medical domain; \citet{wu-cotfails-2025a} showed that CoT prompting
can hurt clinical-task accuracy. Our work complements these by providing a
rigorous quantification of the contamination artifact produced by hindsight-guided
distillation, at substantially lower inference cost than the teacher.

\paragraph{SFT vs.\ reward-based training.}
DeepSeek-R1~\cite{deepseek-r1-2025} showed GRPO-based RL develops genuine reasoning.
\citet{ouyang-instructgpt-2022} established supervised fine-tuning from human feedback
as a foundation for instruction-following; \citet{rafailov-dpo-2023} proposed Direct
Preference Optimisation as a simpler reward-based alternative.
\citet{sft-or-rl-2025} compared SFT and RL for training R1-like reasoning models;
\citet{turpin-unfaithful-2023} identified unfaithful CoT explanations as a signature
of format imitation, and \citet{yang-emperors-2025} formalised this as format
alignment dominating SFT gains. \citet{yang-srft-2025} proposed single-stage SFT+RL
integration, finding that SFT alone produces coarse-grained global changes without
the fine-grained selective optimisations of RL. Our results are consistent with
these characterisations in a clinical diagnostic setting.

%==============================================================================
\section{ZebraMap}
\label{sec:zebramap}

ZebraMap~\cite{zebramap2026} is a multimodal rare-disease knowledge map built from
\ConstZebraMapDiseases{} Orphanet diseases and \ConstZebraMapPubMedCases{}
PubMed case report articles (yielding \ConstZebraMapPatientCases{} patient-level
cases with standardised fields), along with 98,038 linked clinical images.
Case structuring follows a hybrid pipeline: rule-based section detection and
regex-based demographic extraction first isolate individual patient narratives,
which are then passed to an LLM (Gemini) for standardised field population;
a separate RAG component fills missing disease-level fields from retrieved abstracts.

\paragraph{Splits.}
We define our own train/evaluation split from ZebraMap; no canonical split
is provided by the dataset. Training: \ConstZebraMapTrainDiseases{} diseases,
\ConstZebraMapTrainCases{} cases. Evaluation:
\ConstZebraMapEvalDiseases{} diseases $\times$
\ConstZebraMapEvalCasesPerDisease{} cases $=$ \ConstZebraMapEvalCasesTotal{}
total. All hindsight-guided teacher traces are generated exclusively from
training-split cases; no evaluation case is seen during trace generation or
fine-tuning. All evaluation diseases also appear in the training split
(frequency \ConstZebraMapFreqMin{}--\ConstZebraMapFreqMax{}, median
\ConstZebraMapFreqMedian{}); the analysis therefore characterizes knowledge
transfer at varying training exposure levels, not zero-shot generalization.

%==============================================================================
\section{Methodology}
\label{sec:method}

\subsection{Overview}
\label{sec:overview}

Both models are LLMs. We fine-tune a \ConstStudentParams{} student---starting from
\ConstStudentBase{}-Instruct---on reasoning traces produced by a \ConstTeacherParams{}
teacher. The pipeline has four stages.

\paragraph{(i) Hindsight trace generation.}
For each training case, the teacher receives the patient case narrative \emph{and}
the ground-truth disease label, and is instructed to reserve explicit label use for
the end of its reasoning. It emits a \texttt{<think>}\ldots\texttt{</think>} chain
followed by \texttt{<diagnosis>Name</diagnosis>}.

\paragraph{(ii) Redaction.}
We build two training sets from the same traces. For \textbf{StudentF}, a regex
filter deletes every explicit ``ground truth is X'' slot (the phrase and its
associated label mention); for \textbf{Student}, the traces are left untouched.

\paragraph{(iii) Supervised fine-tuning.}
Student and StudentF share every hyperparameter and differ only in their training
traces. Each example is a (narrative, trace) pair, and the loss is standard
next-token cross-entropy over the full trace (\texttt{<think>} plus
\texttt{<diagnosis>}) conditioned on the narrative alone. We use QLoRA
($r{=}\ConstLoRArank{}$, $\alpha{=}\ConstLoRAalpha{}$, 4-bit \ConstStudentQuant{})
for 3 epochs with batch size 32 (4$\times$8 gradient accumulation) and learning
rate $2{\times}10^{-4}$ cosine; full settings are in Appendix~\ref{app:reproducibility}.

\paragraph{(iv) Inference.}
Both students receive the case narrative alone---with no label access---and are
scored on the content of their \texttt{<diagnosis>} tag. The teacher is evaluated
zero-shot on the same narrative-only prompts. All teacher traces are generated from
training-split cases; no evaluation case is seen during trace generation or
fine-tuning.

\subsection{Models}

\paragraph{Base.}
\ConstStudentBase{}-Instruct~\cite{qwen25-2024} (4-bit \ConstStudentQuant{}).
Evaluated in two modes: free-form prose (Mentioned baseline) and zero-shot
\texttt{<diagnosis>}-tag prompting with no chain-of-thought (Accuracy reference
baseline; labelled \emph{Base (tagged)} in tables).

\paragraph{Teacher.}
\ConstTeacherBase{}~\cite{deepseek-r1-2025} (GGUF \ConstTeacherQuant{}).
A CoT distillate of DeepSeek-R1-671B (GRPO-trained); evaluated zero-shot on ZebraMap.

\paragraph{StudentF (filtered; primary model).}
\ConstStudentBase{} fine-tuned with QLoRA~\cite{dettmers-qlora-2023} (rank \ConstLoRArank{},
$\alpha{=}\ConstLoRAalpha{}$, BnB 4-bit) on \ConstZebraMapTrainCases{} teacher-generated
traces from which a regex filter has removed all explicit ``ground truth is X'' slots.
StudentF produces near-zero GT phrases at inference (3/\ConstZebraMapEvalCasesTotal{} = \StatSFGTFoundRate{} contamination rate).

\paragraph{Student (unfiltered; ablation).}
Identical architecture and training to StudentF, but trained on raw teacher traces.
Presented alongside StudentF to reveal the structure and magnitude of GT hallucination.

\subsection{Hindsight Trace Generation}
\label{sec:traces}

Each trace is generated by prompting the teacher with a case narrative \emph{and}
the ground-truth disease label, instructing it to reserve explicit label use for the
end of its reasoning. The teacher outputs \texttt{<think>}\ldots\texttt{</think>}
followed by \texttt{<diagnosis>Name</diagnosis>}. At inference, both Student and
StudentF receive only the case narrative.

This produces label-anchored evidence chains at training time. Because
\ConstTeacherBase{} writes ``ground truth is X'' inside its transparent deliberative
reasoning~\cite{deepseek-r1-2025}, SFT copies this pattern into both trained
models. This is a known trace contamination mechanism~\cite{green-leaky-2025,turpin-unfaithful-2023}:
privileged information present at generation time leaks into outputs through
pattern-copying. StudentF's filter removes these slots from all
\ConstZebraMapTrainCases{} training traces, so the pattern is near-eliminated
during its fine-tuning.

\paragraph{Worked example.}
We illustrate the filtering operation on one training case: an 18-year-old with
mild facial and palatal weakness, fatigue of the lower-limb muscles on exertion,
and a mildly elevated creatine kinase (CK), labelled \emph{Nemaline myopathy}.

\textbf{Input (teacher and student).} The case narrative (abridged):
\begin{quote}\small\itshape
The proband, an 18-year-old \dots\ presented at 8 years of age with mild facial and
palatal weakness and minor fatigue of his lower extremity muscles when swimming
\dots\ Examination showed \dots\ myopathic facies, mild quadriceps and ankle
dorsiflexion weakness \dots\ Creatine kinase (CK) was mildly elevated at 249
(reference 30--150 IU/L).
\end{quote}

\textbf{Raw teacher target} (generated with the label visible):
\begin{quote}\small\itshape
\dots these are classic signs of nemaline myopathy \dots The CK being slightly
elevated is also consistent with this condition. \textbf{I think the ground truth
is \emph{Nemaline myopathy}, which aligns with all the provided information.} The
patient's symptoms and test results fit well within this diagnosis.
\texttt{</think> <diagnosis>Nemaline myopathy</diagnosis>}
\end{quote}

\textbf{StudentF target} (after regex filtering). The bolded sentence is deleted;
every other token is unchanged:
\begin{quote}\small\itshape
\dots these are classic signs of nemaline myopathy \dots The CK being slightly
elevated is also consistent with this condition. The patient's symptoms and test
results fit well within this diagnosis.
\texttt{</think> <diagnosis>Nemaline myopathy</diagnosis>}
\end{quote}

At inference, both students see the narrative only---the label and the entire
``ground truth'' sentence are absent---and are scored on the \texttt{<diagnosis>}
content.

\subsection{Evaluation Metrics}
\label{sec:metrics}

We define four metrics, all novel for this evaluation setting. M1 and M3 are
shorthand labels used in tables and figures throughout.

\textbf{Mentioned (M1)} checks whether the ground-truth disease name appears
anywhere in the full response. It is a search-space proxy only and is never
used as an accuracy measure; it credits passing mentions in rejected hypotheses
as much as committed diagnoses.

\textbf{Accuracy (M3)} is the primary accuracy metric. It extracts the content
of the \texttt{<diagnosis>} tag and scores it by bidirectional substring match
against a disease synonym list; responses without a tag score~0. Only the
model's committed answer is credited. The Base model is also evaluated in a
zero-shot tag-prompting mode with no chain-of-thought, referred to as
\emph{Base (tagged)} in Table~\ref{tab:main}.

\textbf{Accuracy, upper bound (M3\_posttag)} is a robustness bound for the
teacher. For untagged responses it searches post-think prose; for tagged
responses it is identical to M3. This corrects for zero-shot format
non-compliance and represents the best-case Accuracy for teacher evaluation.

\textbf{Accuracy, format-compliant (M3\_restricted)} is computed only over
the $n{=}\StatNRestricted{}$ responses containing a well-formed
\texttt{<diagnosis>} tag, removing any confound from format non-compliance.

\paragraph{Statistical tests.}
Primary: clustered permutation (\ConstZebraMapEvalDiseases{} disease clusters,
$B{=}\ConstPermB{}$). Confidence intervals: 95\% clustered bootstrap
($B{=}\ConstBootstrapB{}$). Holm--Bonferroni correction within each comparison
family.

%==============================================================================
\section{Results}
\label{sec:results}

\subsection{Main Accuracy}

\begin{table*}[tp]
\centering
\small % ACL prefers \small or \footnotesize over \scriptsize if possible; change back if space is tight
\setlength{\tabcolsep}{6pt}
\renewcommand{\arraystretch}{1.2}
\begin{tabularx}{\textwidth}{>{\raggedright\arraybackslash}Xcccc}
\toprule
\textbf{Metric} & \textbf{Base} & \textbf{Student} & \textbf{StudentF} & \textbf{Teacher} \\
\midrule
Mentioned (proxy only)
  & \begin{tabular}[c]{@{}c@{}}\StatBaseMone{}\\ \small[\StatBaseMoneLoCI{},\,\StatBaseMoneHiCI{}]\end{tabular}
  & \begin{tabular}[c]{@{}c@{}}\StatStudMone{}\\ \small[\StatStudMoneLoCI{},\,\StatStudMoneHiCI{}]\end{tabular}
  & \begin{tabular}[c]{@{}c@{}}\StatStudFMone{}\\ \small[\StatStudFMoneLoCI{},\,\StatStudFMoneHiCI{}]\end{tabular}
  & \begin{tabular}[c]{@{}c@{}}\StatTeachMone{}\\ \small[\StatTeachMoneLoCI{},\,\StatTeachMoneHiCI{}]\end{tabular} \\
\addlinespace
\textbf{Accuracy} (primary)
  & \begin{tabular}[c]{@{}c@{}}\StatTagsBidirMthree{}\\ \small[\StatTagsBidirMthreeLoCI{},\,\StatTagsBidirMthreeHiCI{}]\end{tabular}
  & \begin{tabular}[c]{@{}c@{}}\StatStudMthree{}\\ \small[\StatStudMthreeLoCI{},\,\StatStudMthreeHiCI{}]\end{tabular}
  & \begin{tabular}[c]{@{}c@{}}\textbf{\StatStudFMthree{}}\\ \small[\StatStudFMthreeLoCI{},\,\StatStudFMthreeHiCI{}]\end{tabular}
  & \begin{tabular}[c]{@{}c@{}}\StatTeachMthree{}\\ \small[\StatTeachMthreeLoCI{},\,\StatTeachMthreeHiCI{}]\end{tabular} \\
\addlinespace
Accuracy, upper bound
  & ---
  & \begin{tabular}[c]{@{}c@{}}\StatStudMthreePT{}\\ \small[\StatStudMthreePTLo{},\,\StatStudMthreePTHi{}]\end{tabular}
  & \begin{tabular}[c]{@{}c@{}}\textbf{\StatStudFMthreePT{}}\\ \small[\StatStudFMthreePTLo{},\,\StatStudFMthreePTHi{}]\end{tabular}
  & \begin{tabular}[c]{@{}c@{}}\StatTeachMthreePT{}\\ \small[\StatTeachMthreePTLo{},\,\StatTeachMthreePTHi{}]\end{tabular} \\
\addlinespace
Accuracy, format-compliant ($n=\StatNRestricted{}$)
  & ---
  & \begin{tabular}[c]{@{}c@{}}\StatStudMthreeRestr{}\\ \small[\StatStudMthreeRestrLo{},\,\StatStudMthreeRestrHi{}]\end{tabular}
  & \begin{tabular}[c]{@{}c@{}}\textbf{\StatStudFMthreeRestr{}}\\ \small[\StatStudFMthreeRestrLo{},\,\StatStudFMthreeRestrHi{}]\end{tabular}
  & \begin{tabular}[c]{@{}c@{}}\StatTeachMthreeRestr{}\\ \small[\StatTeachMthreeRestrLo{},\,\StatTeachMthreeRestrHi{}]\end{tabular} \\
\midrule
\texttt{<diagnosis>} present
  & --- & \StatFormatSHasDiagRate{} & \StatFormatSFHasDiagRate{} & \StatFormatTHasDiagRate{} \\
Mean response length (chars)
  & \StatMeanRespLenBase{} & \StatMeanRespLenStud{} & \StatMeanRespLenStudF{} & \StatMeanRespLenTeach{} \\
\bottomrule
\end{tabularx}
\caption{All metrics for \ConstZebraMapEvalCasesTotal{} cases across \ConstZebraMapEvalDiseases{} diseases. 95\% clustered bootstrap CIs shown below each point estimate. \textbf{Accuracy is the primary metric.} Mentioned is a search-space proxy and is never used as an accuracy measure. Base (tagged): base model in zero-shot tag-prompting mode, no chain-of-thought.}
\label{tab:main}
\end{table*}

Table~\ref{tab:main} reports all metrics with 95\% clustered bootstrap CIs. The
primary result is that StudentF achieves Accuracy\,=\,\StatStudFMthree{}, significantly
exceeding the teacher (\StatTeachMthree{}, $p{<}0.001$) and the unfiltered Student
(\StatStudMthree{}, $p{<}0.05$). Critically, the unfiltered Student alone does
\emph{not} significantly outperform the teacher ($p{=}\StatPAdjStudVsTeachMthree{}$; Table~\ref{tab:pairwise}),
demonstrating that filtering---not hindsight distillation per se---drives the accuracy gain.
The contamination pattern underlying these comparisons is a known consequence of
label-visible trace generation~\cite{green-leaky-2025}; our contribution is to quantify
its prevalence, per-stratum accuracy impact, and the net effect of filtering with
statistical rigor.

The teacher's Mentioned (\StatTeachMone{}) also
substantially exceeds its Accuracy (\StatTeachMthree{}) by
\StatGapTeachMoneVsMthree{} because it frequently names the correct disease
in deliberation without placing it in a \texttt{<diagnosis>} tag
(\StatFormatTHasDiagRate{} tag compliance). Accuracy (upper bound)
(\StatTeachMthreePT{}) partially recovers these; even so, StudentF's Accuracy
(upper bound) (\StatStudFMthreePT{}) exceeds the teacher's upper bound.

\paragraph{Consistency of the filtered advantage.}
Three alternative metrics confirm the advantage does not depend on the primary
Accuracy metric's tag requirement.
On Accuracy (format-compliant) ($n{=}\StatNRestricted{}$ format-compliant cases): StudentF
\StatStudFMthreeRestr{} vs.\ teacher \StatTeachMthreeRestr{}
($p{<}0.05$). On Accuracy (upper bound): StudentF \StatStudFMthreePT{} vs.\
teacher \StatTeachMthreePT{} ($p{<}0.05$). All pairwise comparisons
are summarised in Table~\ref{tab:pairwise}.

\begin{table*}[t]
\small % Compliant size; use \footnotesize if you need exactly 9pt.
\centering
\begin{tabularx}{\textwidth}{@{}X l c c@{}}
\toprule
\textbf{Comparison} & \textbf{Metric} & \textbf{RD [95\% CI]} & \textbf{Cohen's $h$ [95\% CI]} \\
\midrule
\multicolumn{4}{l}{\textit{Primary (M3/Accuracy)}} \\
\quad StudentF vs.\ Teacher & Acc. & $\StatEffStudFVsTeachMthreeRD{}$ [$\StatEffStudFVsTeachMthreeRDLo{}$, $\StatEffStudFVsTeachMthreeRDHi{}$] & $\StatEffStudFVsTeachMthreeH{}$ [$\StatEffStudFVsTeachMthreeHLo{}$, $\StatEffStudFVsTeachMthreeHHi{}$] \\
\quad Student vs.\ Teacher & Acc. & $\StatEffStudVsTeachMthreeRD{}$ [$\StatEffStudVsTeachMthreeRDLo{}$, $\StatEffStudVsTeachMthreeRDHi{}$] & $\StatEffStudVsTeachMthreeH{}$ [$\StatEffStudVsTeachMthreeHLo{}$, $\StatEffStudVsTeachMthreeHHi{}$] \\
\quad StudentF vs.\ Student & Acc. & $\StatPairMthreeStudentVsStudentFRD{}$ [$\StatPairMthreeStudentVsStudentFRDLo{}$, $\StatPairMthreeStudentVsStudentFRDHi{}$] & $\StatPairMthreeStudentVsStudentFH{}$ [$\StatPairMthreeStudentVsStudentFHLo{}$, $\StatPairMthreeStudentVsStudentFHHi{}$] \\
\quad StudentF vs.\ Base & Acc. & $\StatPairMthreeTagsVsStudentFRD{}$ [$\StatPairMthreeTagsVsStudentFRDLo{}$, $\StatPairMthreeTagsVsStudentFRDHi{}$] & $\StatPairMthreeTagsVsStudentFH{}$ [$\StatPairMthreeTagsVsStudentFHLo{}$, $\StatPairMthreeTagsVsStudentFHHi{}$] \\
\midrule
\multicolumn{4}{l}{\textit{Conservative bounds}} \\
\quad StudentF vs.\ Teacher & Acc.\ (UB) & $\StatPairMthreePTStudentFVsTeacherRD{}$ [$\StatPairMthreePTStudentFVsTeacherRDLo{}$, $\StatPairMthreePTStudentFVsTeacherRDHi{}$] & $\StatPairMthreePTStudentFVsTeacherH{}$ [$\StatPairMthreePTStudentFVsTeacherHLo{}$, $\StatPairMthreePTStudentFVsTeacherHHi{}$] \\
\quad StudentF vs.\ Teacher & Acc.\ (FC) & $\StatPairMthreeRStudentFVsTeacherRD{}$ [$\StatPairMthreeRStudentFVsTeacherRDLo{}$, $\StatPairMthreeRStudentFVsTeacherRDHi{}$] & $\StatPairMthreeRStudentFVsTeacherH{}$ [$\StatPairMthreeRStudentFVsTeacherHLo{}$, $\StatPairMthreeRStudentFVsTeacherHHi{}$] \\
\bottomrule
\end{tabularx}
\caption{Pairwise comparisons. RD = risk difference (A--B); Cohen's $h$ = effect size on arcsine scale. All CIs: 95\% clustered bootstrap; $p$-values: Holm--Bonferroni adjusted clustered permutation. StudentF exceeds the teacher under every Accuracy variant ($p < 0.001$ primary; $p < 0.01$ conservative bounds). Student alone does not significantly outperform the teacher ($p = \StatPAdjStudVsTeachMthree{}$). StudentF vs.\ Student ($p = \StatPAdjStudVsStudFMthree{}$) is significant, confirming that filtering---not distillation alone---drives the gain; see Section~\ref{sec:gt_hallucination}.}
\label{tab:pairwise}
\end{table*}

\subsection{Quantifying Trace Contamination}
\label{sec:gt_hallucination}

\begin{table*}[tp]
\small\centering
\setlength{\tabcolsep}{4pt}
\renewcommand{\arraystretch}{1.2}
\begin{tabular}{@{}lcccc@{}}
\toprule
\textbf{Stratum (defined by Student's reasoning chain)} &
  $n$ & \textbf{Student Accuracy} & \textbf{StudentF Accuracy} & \textbf{Teacher Accuracy} \\
\midrule
No GT phrase
  & \ConstGTAbsentStratumN{}
  & \makecell[c]{\StatStudMthreeNoGT{}\\\small[\StatStudMthreeNoGTLoCI{},\,\StatStudMthreeNoGTHiCI{}]}
  & \makecell[c]{\StatSFMthreeNoGT{}\\\small[\StatSFMthreeNoGTLoCI{},\,\StatSFMthreeNoGTHiCI{}]}
  & \makecell[c]{\StatTeachMthreeNoGT{}\\\small[\StatTeachMthreeNoGTLoCI{},\,\StatTeachMthreeNoGTHiCI{}]} \\
GT phrase, label correct
  & \ConstGTCorrectStratumN{}
  & \makecell[c]{\StatStudMthreeGTright{}\\\small[\StatStudMthreeGTrightLoCI{},\,\StatStudMthreeGTrightHiCI{}]}
  & \makecell[c]{\StatSFMthreeGTright{}\\\small[\StatSFMthreeGTrightLoCI{},\,\StatSFMthreeGTrightHiCI{}]}
  & \makecell[c]{\StatTeachMthreeGTright{}\\\small[\StatTeachMthreeGTrightLoCI{},\,\StatTeachMthreeGTrightHiCI{}]} \\
GT phrase, label wrong
  & \ConstGTWrongStratumN{}
  & \makecell[c]{\StatStudMthreeGTwrong{}\\\small[\StatStudMthreeGTwrongLoCI{},\,\StatStudMthreeGTwrongHiCI{}]}
  & \makecell[c]{\StatSFMthreeGTwrong{}\\\small[\StatSFMthreeGTwrongLoCI{},\,\StatSFMthreeGTwrongHiCI{}]}
  & \makecell[c]{\StatTeachMthreeGTwrong{}\\\small[\StatTeachMthreeGTwrongLoCI{},\,\StatTeachMthreeGTwrongHiCI{}]} \\
\bottomrule
\end{tabular}
\caption{Accuracy stratified by GT phrase presence in the Student's reasoning
chain. Strata are defined by Student behaviour; StudentF and Teacher values cover
the same case subsets for comparability. StudentF produces near-zero GT phrases
at inference (\StatNGTFoundStudF{}/\ConstZebraMapEvalCasesTotal{}); Teacher
produces none. Because strata are defined retrospectively by Student's own outputs
and the models are unmatched on these cases, comparisons within strata are
descriptive, not causal. Clustered permutation test (Student: GT-wrong vs.\
GT-absent): $p{<}0.001$.}
\label{tab:gt}
\end{table*}

Trace contamination from label-visible generation is a known
mechanism~\cite{green-leaky-2025,turpin-unfaithful-2023}. Here it takes the form of
``ground truth is X'' phrases: the teacher writes them during trace generation
because the label is part of its input, and SFT copies the pattern. Student
reproduces it at inference despite no label access; StudentF does not, because
the pattern was near-eliminated from its training data.

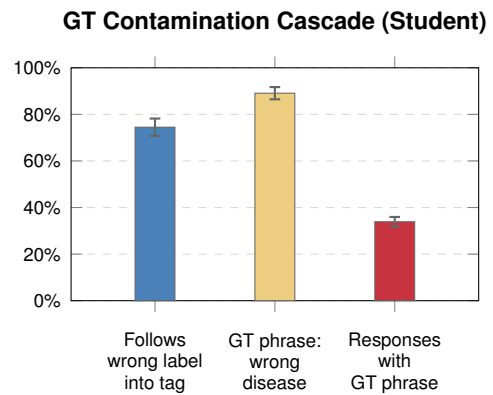
\begin{figure}[htb]
\centering
\begin{tikzpicture}[trim axis left]
\begin{axis}[
    ybar,
    width=0.7\columnwidth,    
    height=0.4\columnwidth,   % Slightly increased height to balance the larger text
    ymin=0, ymax=1,
    ytick={0,0.2,0.4,0.6,0.8,1.0},
    yticklabels={0\%,20\%,40\%,60\%,80\%,100\%},
    yticklabel style={font=\scriptsize\sffamily}, % Increased from \tiny to \scriptsize
    xtick={0,1,2},
    xticklabels={%
      {Follows\\wrong label\\into tag}, % Added (n=...) placeholder
      {GT phrase:\\wrong\\disease},     % Added (n=...) placeholder
      {Responses\\with\\GT phrase}},    % Added (n=...) placeholder
    xticklabel style={font=\scriptsize\sffamily, align=center, yshift=-4pt}, % Increased font size and added comfort padding
    bar width=15pt,            
    enlarge x limits=0.35,     % Tweaked to give the larger text blocks maximum breathing room
    ymajorgrids=true,
    grid style={dashed,gray!30},
    title={GT Contamination Cascade (Student)},
    title style={font=\small\bfseries\sffamily, yshift=3pt},
    scale only axis,
    clip=false,
]

\addplot[
    fill=teal_color, draw=black!60, bar shift=0pt,
    error bars/.cd,
        y dir=both, y explicit,
        error bar style={thick, black!60},
        error mark options={rotate=90, mark size=2pt, thick}
] coordinates {
    (0, \StatGTFollowsRate) +- (0, \fpeval{\StatGTFollowsRate - \StatGTFollowsLoCI})
};

\addplot[
    fill=gold_color, draw=black!60, bar shift=0pt,
    error bars/.cd,
        y dir=both, y explicit,
        error bar style={thick, black!60},
        error mark options={rotate=90, mark size=2pt, thick}
] coordinates {
    (1, \StatGTWrongCondRate) +- (0, \fpeval{\StatGTWrongCondRate - \StatGTWrongCondLoCI})
};

\addplot[
    fill=red_color, draw=black!60, bar shift=0pt,
    error bars/.cd,
        y dir=both, y explicit,
        error bar style={thick, black!60},
        error mark options={rotate=90, mark size=2pt, thick}
] coordinates {
    (2, \StatGTFoundRate) +- (0, \fpeval{\StatGTFoundRate - \StatGTFoundLoCI})
};

\end{axis}
\end{tikzpicture}
\caption{GT contamination cascade for the unfiltered Student.
Of \ConstZebraMapEvalCasesTotal{} inferences,
\fpeval{round(\StatGTFoundRate*100,1)}\%
[\StatGTFoundLoCI{}, \StatGTFoundHiCI{}] contain a GT phrase;
of those, \fpeval{round(\StatGTWrongCondRate*100,1)}\%
[\StatGTWrongCondLoCI{}, \StatGTWrongCondHiCI{}] name the wrong disease;
of those, \fpeval{round(\StatGTFollowsRate*100,1)}\%
[\StatGTFollowsLoCI{}, \StatGTFollowsHiCI{}] are copied into the
\texttt{<diagnosis>} tag. Error bars: 95\% clustered bootstrap CIs.}
\label{fig:cascade}
\end{figure}

\begin{figure}[htb]
\centering
\begin{tikzpicture}
\begin{axis}[
    width=0.7\columnwidth,
    height=0.4\columnwidth,
    ybar=3pt,
    bar width=9pt,
    ymin=0, ymax=1.0,
    ylabel={Accuracy (M3)},
    ylabel style={font=\small\sffamily},
    symbolic x coords={NoGT,GTcorrect,GTwrong},
    xtick=data,
    xticklabel style={align=center, font=\small\sffamily},
    xticklabels={%
      No GT\\in think\\(n=\ConstGTAbsentStratumN),
      GT correct\\label\\(n=\ConstGTCorrectStratumN),
      GT wrong\\label\\(n=\ConstGTWrongStratumN)},
    enlarge x limits=0.25,
    xtick align=outside,
    extra x tick style={tick style={draw=none}},
    % Leave room below axis for n= labels
    xticklabel shift=10pt,
    legend style={
        at={(0.5,-0.4)},
        anchor=north,
        legend columns=3,
        draw=none,
        font=\small\sffamily,
        column sep=6pt,
        inner sep=3pt,
    },
    legend image post style={scale=0.8},
    ymajorgrids=true,
    grid style={dashed,gray!30},
    title={Accuracy by GT Stratum (95\% bootstrap CIs)},
    title style={font=\small\bfseries\sffamily, yshift=4pt},
    scale only axis,
    clip=false,
]

\addplot[
    fill=teal_color, draw=black!70,
    error bars/.cd, y dir=both, y explicit,
    error bar style={thick,black!70},
    error mark options={rotate=90,mark size=2pt,thick}
] coordinates {
    (NoGT,\StatStudMthreeNoGT)
        +-(0,\fpeval{(\StatStudMthreeNoGTHiCI-\StatStudMthreeNoGTLoCI)/2})
    (GTcorrect,\StatStudMthreeGTright)
        +-(0,\fpeval{(\StatStudMthreeGTrightHiCI-\StatStudMthreeGTrightLoCI)/2})
    (GTwrong,\StatStudMthreeGTwrong)
        +-(0,\fpeval{(\StatStudMthreeGTwrongHiCI-\StatStudMthreeGTwrongLoCI)/2})
};
\addplot[
    fill=gold_color, draw=black!70,
    error bars/.cd, y dir=both, y explicit,
    error bar style={thick,black!70},
    error mark options={rotate=90,mark size=2pt,thick}
] coordinates {
    (NoGT,\StatSFMthreeNoGT)
        +-(0,\fpeval{(\StatSFMthreeNoGTHiCI-\StatSFMthreeNoGTLoCI)/2})
    (GTcorrect,\StatSFMthreeGTright)
        +-(0,\fpeval{(\StatSFMthreeGTrightHiCI-\StatSFMthreeGTrightLoCI)/2})
    (GTwrong,\StatSFMthreeGTwrong)
        +-(0,\fpeval{(\StatSFMthreeGTwrongHiCI-\StatSFMthreeGTwrongLoCI)/2})
};
\addplot[
    fill=red_color, draw=black!70,
    error bars/.cd, y dir=both, y explicit,
    error bar style={thick,black!70},
    error mark options={rotate=90,mark size=2pt,thick}
] coordinates {
    (NoGT,\StatTeachMthreeNoGT)
        +-(0,\fpeval{(\StatTeachMthreeNoGTHiCI-\StatTeachMthreeNoGTLoCI)/2})
    (GTcorrect,\StatTeachMthreeGTright)
        +-(0,\fpeval{(\StatTeachMthreeGTrightHiCI-\StatTeachMthreeGTrightLoCI)/2})
    (GTwrong,\StatTeachMthreeGTwrong)
        +-(0,\fpeval{(\StatTeachMthreeGTwrongHiCI-\StatTeachMthreeGTwrongLoCI)/2})
};

\legend{Student, StudentF, Teacher}

\end{axis}
\end{tikzpicture}
\caption{M3 accuracy stratified by GT phrase presence in Student's reasoning chain.
StudentF: zero GT phrases at inference (filter removed all from training).
Teacher: zero GT phrases (zero-shot; no label access).
Strata are defined by Student's own reasoning, so cross-model comparisons within a
stratum are descriptive rather than causal.
Clustered permutation, Student GT-wrong vs.\ GT-absent: $p{<}0.001$.}
\label{fig:main}
\end{figure}
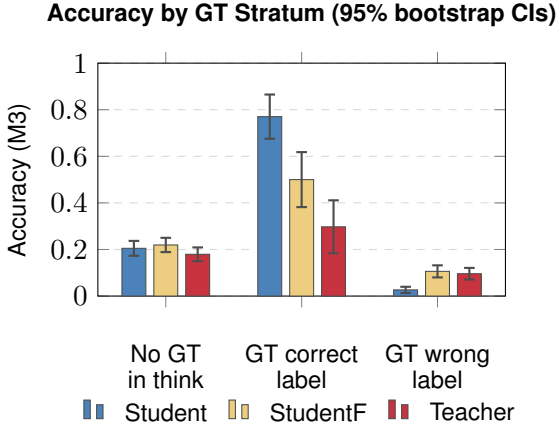

\paragraph{Contamination cascade.}
A GT phrase appears in
\fpeval{round(\StatGTFoundRate*100,1)}\% [\StatGTFoundLoCI{}, \StatGTFoundHiCI{}]
of Student inferences. Of those, \fpeval{round(\StatGTWrongCondRate*100,1)}\%
[\StatGTWrongCondLoCI{}, \StatGTWrongCondHiCI{}] contain a wrong disease. When
the wrong label is present, Student copies it into \texttt{<diagnosis>}
\fpeval{round(\StatGTFollowsRate*100,1)}\% [\StatGTFollowsLoCI{}, \StatGTFollowsHiCI{}]
of the time. Table~\ref{tab:gt} shows the resulting Accuracy stratification: Student
drops from \StatStudMthreeNoGT{} (GT absent) to \StatStudMthreeGTwrong{} (GT wrong),
a \fpeval{round((\StatStudMthreeNoGT - \StatStudMthreeGTwrong)*100, 0)} percentage
point collapse. Figures~\ref{fig:cascade} and~\ref{fig:main} illustrate the cascade
proportions and the resulting per-stratum Accuracy.

\paragraph{Why the filtering gain is bounded.}
StudentF suppresses GT contamination to near-zero at inference
(\StatNGTFoundStudF{}/\ConstZebraMapEvalCasesTotal{} = \StatSFGTFoundRate{}),
gaining \fpeval{round(\StatSFMthreeGTwrong - \StatStudMthreeGTwrong, 3)} Accuracy over
Student on the GT-wrong stratum---the intended benefit. The GT-correct stratum
($n{=}\ConstGTCorrectStratumN{}$) is descriptively informative: Student Accuracy
there (\StatStudMthreeGTright{}) is far above its own uncontaminated baseline
(\StatStudMthreeNoGT{}), which is consistent with correct GT hints contributing
diagnostic value when they happen to be followed. StudentF, which never sees the
phrase, scores \StatSFMthreeGTright{} on this stratum ($-$\fpeval{round(\StatStudMthreeGTright - \StatSFMthreeGTright, 3)} Accuracy relative to Student). We caution against a causal reading of this gap: the
stratum is defined post hoc by Student's own inference behaviour, the two models
are not matched on these cases before treatment, and cases in which Student
spontaneously emits a correct GT phrase are plausibly easier, so part of the gap
may reflect case difficulty rather than the removal of a correct hint. The net
effect remains a \StatStudFVsStudMthreeGap{} Accuracy gain for StudentF overall
($p{<}0.05$), indicating that filtering is beneficial on balance.

Crucially, the advantage is not limited to contamination removal: on the \StatStudMthreeNoGT{}
GT-absent stratum ($n{=}\ConstGTAbsentStratumN{}$), StudentF (\StatSFMthreeNoGT{}) also
significantly outperforms the Teacher (\StatTeachMthreeNoGT{}, $p_\text{adj}{=}0.009$),
while Student does not ($p_\text{adj}{=}0.17$). Filtering therefore improves not only
the contaminated cases but also produces better clean-case reasoning than
unfiltered SFT.

Aggressive filtering is also inherently all-or-nothing. Training-time labels are
always correct, so a static filter cannot know in advance which phrases a student
would have reproduced correctly at inference. What the data establish is that
removing all such phrases yields a positive net effect; whether a more selective
filter could recover additional gains remains open.

%==============================================================================
\section{Analysis}
\label{sec:analysis}

\subsection{Training Frequency and Knowledge Transfer}
\label{sec:freq}

\begin{table}[tbp]
\small\centering
\setlength{\tabcolsep}{3pt}
\renewcommand{\arraystretch}{1.15}
\begin{tabular}{@{}lccccc@{}}
\toprule
\textbf{Q} & \textbf{Freq} &
  \textbf{Stud Acc.} & \textbf{SF Acc.} & \textbf{Teach Acc.} \\
\midrule
Q1 & \ConstZebraMapFreqQoneMedian{}
  & \makecell[c]{\StatFreqQoneStudMthree{}\\\small[\StatFreqQoneStudMthreeLoCI{},\,\StatFreqQoneStudMthreeHiCI{}]}
  & \makecell[c]{\StatFreqQoneStudFMthree{}\\\small[\StatFreqQoneStudFMthreeLoCI{},\,\StatFreqQoneStudFMthreeHiCI{}]}
  & \makecell[c]{\StatFreqQoneTeachMthree{}\\\small[\StatFreqQoneTeachMthreeLoCI{},\,\StatFreqQoneTeachMthreeHiCI{}]}\\
Q2 & \ConstZebraMapFreqQtwoMedian{}
  & \makecell[c]{\StatFreqQtwoStudMthree{}\\\small[\StatFreqQtwoStudMthreeLoCI{},\,\StatFreqQtwoStudMthreeHiCI{}]}
  & \makecell[c]{\StatFreqQtwoStudFMthree{}\\\small[\StatFreqQtwoStudFMthreeLoCI{},\,\StatFreqQtwoStudFMthreeHiCI{}]}
  & \makecell[c]{\StatFreqQtwoTeachMthree{}\\\small[\StatFreqQtwoTeachMthreeLoCI{},\,\StatFreqQtwoTeachMthreeHiCI{}]}\\
Q3 & \ConstZebraMapFreqQthreeMedian{}
  & \makecell[c]{\StatFreqQthreeStudMthree{}\\\small[\StatFreqQthreeStudMthreeLoCI{},\,\StatFreqQthreeStudMthreeHiCI{}]}
  & \makecell[c]{\StatFreqQthreeStudFMthree{}\\\small[\StatFreqQthreeStudFMthreeLoCI{},\,\StatFreqQthreeStudFMthreeHiCI{}]}
  & \makecell[c]{\StatFreqQthreeTeachMthree{}\\\small[\StatFreqQthreeTeachMthreeLoCI{},\,\StatFreqQthreeTeachMthreeHiCI{}]}\\
Q4 & \ConstZebraMapFreqQfourMedian{}
  & \makecell[c]{\textbf{\StatFreqQfourStudMthree{}}\\\small[\StatFreqQfourStudMthreeLoCI{},\,\StatFreqQfourStudMthreeHiCI{}]}
  & \makecell[c]{\textbf{\StatFreqQfourStudFMthree{}}\\\small[\StatFreqQfourStudFMthreeLoCI{},\,\StatFreqQfourStudFMthreeHiCI{}]}
  & \makecell[c]{\StatFreqQfourTeachMthree{}\\\small[\StatFreqQfourTeachMthreeLoCI{},\,\StatFreqQfourTeachMthreeHiCI{}]}\\
\midrule
$\rho$ &&
  \StatRhoFreqVsSMthree{}$^{**}$ &
  \StatRhoFreqVsSFMthree{}$^{*}$ &
  \StatRhoFreqVsTMthree{}~(ns) & \\
OLS &&
  \StatOLSSMthreeSlope{}$^{**}$ &
  \StatOLSSFMthreeSlope{}$^{*}$ &
  \StatOLSTMthreeSlope{}~(ns) & \\
\bottomrule
\end{tabular}
\caption{Mean per-disease Accuracy by training-frequency quartile with 95\% clustered
bootstrap CIs. The StudentF--Teacher gap grows from $\StatFreqSfMinusTQone{}$
in Q1 (where Teacher marginally leads) to $\StatFreqSfMinusTQfour{}$ in Q4.
Teacher shows no frequency dependence.
$^{**}p{<}0.01$, $^{*}p{<}0.05$, ns: $p{\geq}0.05$.}
\label{tab:freq}
\end{table}

StudentF Accuracy scales significantly with training frequency (Spearman
$\rho{=}\StatRhoFreqVsSFMthree{}$, $p{<}0.01$; OLS slope
\StatOLSSFMthreeSlope{}, $p{<}0.05$). The teacher shows no
significant frequency dependence ($\rho{=}\StatRhoFreqVsTMthree{}$,
$p{=}0.65$ (ns)). This dissociation is consistent with the student's
gains reflecting knowledge transfer from training traces rather than model
capacity, though it is also consistent with format alignment and
memorisation of common disease presentations~\cite{yang-emperors-2025,wadhwa-mysteries-2024}.
The StudentF--Teacher gap grows from $\StatFreqSfMinusTQone{}$ in Q1
to \StatFreqSfMinusTQfour{} in Q4 ($\rho$ for SF$-$T gap:
\StatRhoFreqVsSFMinusT{}, $p{<}0.001$). At Q1 (median
\ConstZebraMapFreqQoneMedian{} training cases) Teacher marginally leads StudentF
($\StatFreqSfMinusTQone{}$) and all three models are statistically
indistinguishable---SFT cannot create knowledge absent from training.

\subsection{Reasoning Style and Epistemic Calibration}
\label{sec:calibration}

\begin{table*}[tp]
\small\centering
\setlength{\tabcolsep}{5pt}
\renewcommand{\arraystretch}{1.2}
\begin{tabular}{@{}lccc@{}}
\toprule
\textbf{Metric} & \textbf{Student} & \textbf{StudentF} & \textbf{Teacher} \\
\midrule
Think length (chars)    & \sigfig{\StatStyleThinkLenS} & \sigfig{\StatStyleThinkLenSF} & \sigfig{\StatStyleThinkLenT} \\
Self-corrections        & \sigfig[2]{\StatStyleSelfCorrectS} & \sigfig[2]{\StatStyleSelfCorrectSF} & \sigfig[2]{\StatStyleSelfCorrectT} \\
Expl.\ uncertainty      & 4.53 & 2.12 & 4.09 \\
Confident assertions    & 0.31 & 0.51 & 0.67 \\
\midrule
SC ratio ($\times$/$\checkmark$)
  & \makecell{\StatSCRatioStudAll{}\\\small[\StatSCRatioStudLoCI{},\,\StatSCRatioStudHiCI{}]}
  & \makecell{\StatSCRatioStudFAll{}\\\small[\StatSCRatioStudFLoCI{},\,\StatSCRatioStudFHiCI{}]}
  & \makecell{\StatSCRatioTeach{}\\\small[\StatSCRatioTeachLoCI{},\,\StatSCRatioTeachHiCI{}]} \\
\bottomrule
\end{tabular}
\caption{Mean reasoning-style features (overall) and SC ratio (95\% clustered
bootstrap CI). All models hedge more when wrong; SC ratio $>1$ directionally
for all three, though StudentF's CI $[\StatSCRatioStudFLoCI{}, \StatSCRatioStudFHiCI{}]$
straddles 1.0.}
\label{tab:calibration}
\end{table*}

Table~\ref{tab:calibration} shows reasoning style by model. All three models
think longer, self-correct more, and express more uncertainty when wrong---a
pattern captured by the SC ratio---but the absolute rates differ dramatically
across models.

Filtering has an unexpectedly strong suppressive effect on StudentF's
reasoning behaviour. StudentF self-corrects at 0.15 per response, significantly below the
teacher's 0.28 ($p{<}0.001$) and far below Student's 0.83. Explicit uncertainty follows the same pattern: StudentF 2.12
vs.\ Teacher 4.09 vs.\ Student 4.53. Both effects are consistent with
GT-label contamination causing extra counter-reasoning in Student traces
(``wait---the ground truth says X, but...''); removing that contamination from
training data suppresses the associated hedging behaviour in StudentF.

The most salient Teacher-vs-Student contrast is in confident assertions. Teacher
commits assertively at 0.67 per response vs.\ 0.31 for Student---a 2.1$\times$
difference ($p{<}0.001$), consistent with RL training toward committed,
evaluable answers~\cite{sft-or-rl-2025,turpin-unfaithful-2023}.
StudentF's filtering partially closes this gap (0.51, $p{<}0.001$ vs.\ Teacher),
suggesting that removing GT contamination also removes the conditionally
uncertain reasoning that suppresses assertion confidence. All models show a
modest positive gap in confident assertions on correct vs.\ wrong answers
with similar relative magnitude~\cite{li-confidence-2024}.

The SC ratio is $>1$ for all three models directionally, but StudentF's CI
$[\StatSCRatioStudFLoCI{}, \StatSCRatioStudFHiCI{}]$ straddles 1.0, meaning
the elevation of self-correction on wrong responses is not statistically
distinguishable from noise for StudentF specifically. Student and Teacher
show more reliable miscalibration ($[\StatSCRatioStudLoCI{}, \StatSCRatioStudHiCI{}]$
and $[\StatSCRatioTeachLoCI{}, \StatSCRatioTeachHiCI{}]$ respectively).
This pattern of format imitation without epistemic calibration is consistent
with findings that SFT produces format alignment rather than genuine adaptive
reasoning~\cite{yang-emperors-2025,jiang-betteracc-2026}.

%==============================================================================
\section{Discussion}
\label{sec:discussion}

\paragraph{What hindsight-guided distillation with contamination filtering achieves.}
StudentF demonstrates that a \ConstStudentParams{} model trained via SFT on
\ConstZebraMapTrainCases{} hindsight-guided traces can achieve a narrow accuracy
advantage over its \ConstTeacherParams{} teacher on ZebraMap's rare-disease
diagnosis task (\StatStudFVsTeachMthreeGap{} M3, $p{<}0.001$). Notably, the
unfiltered Student alone does \emph{not} significantly outperform the teacher
($p{=}\StatPAdjStudVsTeachMthree{}$): it is the combination of hindsight
distillation \emph{and} contamination filtering that drives the gain. The
advantage is concentrated in better-represented diseases; at Q1 (median
\ConstZebraMapFreqQoneMedian{} training cases) Teacher marginally leads StudentF
(\StatFreqSfMinusTQone{}), with all three models statistically
indistinguishable, consistent with the student's gains reflecting knowledge
transfer from well-represented training cases (or format alignment with those
patterns) rather than model capacity. Deploying a \ConstStudentParams{} student
at inference costs far less than serving the \ConstTeacherParams{} teacher.

\paragraph{The limits of static filtering.}
Filtering correct GT labels alongside incorrect ones is unavoidable with a static
regex: the filter cannot observe inference-time behaviour. Our stratum analysis
characterizes an apparent cost: on the GT-correct stratum
($n{=}\ConstGTCorrectStratumN{}$) Student reaches \StatStudMthreeGTright{}, far above
its uncontaminated baseline (\StatStudMthreeNoGT{}), whereas StudentF---which never
sees the phrase---reaches \StatSFMthreeGTright{}. Because these strata are defined
retrospectively by Student's own output and the models are not matched on the cases,
this is a descriptive offset against the gains on the GT-wrong stratum rather than a
causal estimate of removing a correct hint. Despite the offset, filtering yields a
statistically significant net gain overall ($p{<}0.05$). More sophisticated
selection approaches---such as training a classifier to predict inference-time
reproduction accuracy, or optimistic consistency
filtering~\cite{wang-scott-2023,wang-selfconsistency-2022}---may reduce the actual
cost further, but the fundamental limitation persists: the filter never observes the
inference error it is trying to prevent.

\paragraph{SFT without reinforcement learning.}
StudentF acquires the teacher's reasoning format
(\StatFormatSFHasThinkDiagRate{} tag adherence) and frequency-sensitive diagnostic
knowledge, but not epistemic calibration: its confident assertions do not track
correctness, and its SC ratio (\StatSCRatioStudFAll{}) remains above one without
genuine uncertainty meaning. This pattern of format imitation without adaptive
calibration is consistent with findings that SFT produces format alignment rather
than genuine reasoning~\cite{yang-emperors-2025}. ~\citet{jiang-betteracc-2026}
further found that medical CoT distillation degrades factual reliability even when
accuracy improves, suggesting that accuracy gains alone do not indicate trustworthy
reasoning. A reward signal that penalises wrong diagnoses at inference would
simultaneously address both failures---directly improving diagnostic accuracy, and
specifically discouraging hallucinated GT phrases (since following a wrong GT label
into \texttt{<diagnosis>} produces a reliably penalised outcome). GRPO-based training
has shown this effect in general reasoning
settings~\cite{deepseek-r1-2025,shao-deepseekmath-2024}; verbal reinforcement
approaches such as Reflexion~\cite{shinn-reflexion-2023} offer a complementary
direction for iterative self-correction without full RL.

\paragraph{Disease-vocabulary confound and the dataset ceiling.}
BM25 retrieval at $k\!=\!1$ achieves \StatBMTwentyFiveBest{} label-classification accuracy on our evaluation set---more than twice the best model's M3 Accuracy (\StatStudFMthree{})---despite using no medical knowledge beyond token counts. Low ROUGE-L overlap between evaluation cases and their BM25 matches (mean \StatBMTwentyFiveRougeLMean{}, median \StatBMTwentyFiveRougeLMedian{}) confirms that BM25 does not succeed via text copying; the signal is disease-vocabulary leakage: cases concerning the same rare disease share distinctive symptom clusters, syndrome names, and genetic markers that act as near-unique identifiers. A concurrent study on ZebraMap by Alkhathlan et al.~\cite{s26113582} independently corroborates this mechanism: using a hybrid retrieval-and-classification pipeline with no LLM reasoning, they find that a class-balanced TF--IDF classifier is their strongest component (MRR~0.4200, Recall@10~0.6279), with BM25 serving as a complementary but individually weaker signal. Their analysis also applies grouped splitting by source article---a control not present in our disease-level split---suggesting that source-article overlap among training diseases may further inflate the BM25 upper bound. Together, these results reinforce that absolute accuracy figures on ZebraMap at this model scale reflect the disease-vocabulary confound as much as diagnostic reasoning; the relative ordering of models, which is the primary object of our analysis, remains the more interpretable quantity.

%==============================================================================
\section{Conclusion}
\label{sec:conclusion}

We studied hindsight-guided rationale distillation for rare-disease diagnosis,
training a \ConstStudentParams{} student on chain-of-thought traces generated by
an \ConstTeacherParams{} teacher with ground-truth label access. Trace
contamination from label-visible generation is a known
mechanism~\cite{green-leaky-2025}; we precisely quantify its
prevalence (\fpeval{round(\StatGTFoundRate*100,1)}\% of Student inferences),
accuracy impact (18pp collapse when the leaked label is wrong), and the trade-offs
of static filtering in this setting. The filtered variant StudentF achieves
near-zero contamination (3/\ConstZebraMapEvalCasesTotal{}) and a narrow accuracy
advantage over the teacher ($p{<}0.001$), while the unfiltered Student does not
significantly outperform the teacher ($p{=}\StatPAdjStudVsTeachMthree{}$).
Filtering yields a statistically significant net gain ($p{<}0.05$), though a
retrospectively-defined correct-label stratum suggests a partially bounding
descriptive cost. These measurements provide a
rigorous characterization of this contamination artifact in a clinical diagnostic
context, connecting to broader concerns about trace privacy and SFT faithfulness.

%==============================================================================
\section*{Limitations}
\label{sec:limitations}

\paragraph{Frequency-distribution sensitivity.}
StudentF's aggregate advantage over the Teacher ($+\StatEffStudFVsTeachMthreeRD{}$
M3) is entirely concentrated in high-exposure diseases: in Q1 (median
\ConstZebraMapFreqQoneMedian{} training cases) Teacher leads StudentF by
\fpeval{abs(\StatFreqSfMinusTQone)*100}\,pp, with all models statistically
indistinguishable. The aggregate result therefore depends on the frequency
distribution of the evaluation set; an evaluation set weighted toward
low-frequency diseases would reduce or reverse the advantage.

\paragraph{Performance ceiling.}
Even the Teacher achieves only \StatTeachMthree{} M3 Accuracy.
The structural ceiling analysis finds that
\fpeval{round(\StatStructuralCeilingFrac*100,0)}\% of eval diseases
(\StatStructuralCeilingN{}/\ConstZebraMapEvalDiseases{}) have max M3\_posttag
below 0.25 across all models. The mean Spearman correlation between per-disease
M3 accuracy rankings, averaged across all $\binom{4}{2}{=}6$ model pairs, is
$\bar{\rho}{=}\StatMeanPairwiseRho{}$---indicating moderate difficulty
concordance across models. This shared difficulty profile suggests the binding
constraint is the task itself at this model scale, not model-specific capacity;
distillation gains accrue on top of a ceiling that affects teacher and students equally.

\paragraph{Trace contamination detection is a lower bound.}
The contamination analysis relies on a single family of regex patterns
targeting ``ground truth is X'' constructions. Alternative phrasings not
captured by this pattern go undetected, so the measured \fpeval{round(\StatGTFoundRate*100,1)}\%
Student contamination rate is a lower bound on true GT exposure in the
reasoning chain.

\paragraph{Static filter cost.}
The regex filter used to produce StudentF removes GT slots without observing
which labels would be reproduced correctly at inference. On the GT-correct
stratum ($n{=}\ConstGTCorrectStratumN{}$), StudentF Accuracy is
\fpeval{round((\StatStudMthreeGTright-\StatSFMthreeGTright)*100,1)}\,pp
lower than Student. Because this stratum is defined retrospectively by Student's
own inference behaviour and the models are not matched on it before treatment,
the gap is a descriptive association rather than a causal estimate of the cost
of excising a correct hint; part of it may reflect that these are intrinsically
easier cases. The structural limitation remains that a static filter cannot
distinguish correct from incorrect labels a priori; closing the gap would
require an inference-time signal.

\paragraph{Pre-distillation baseline.}
We evaluate the Base model only in a zero-shot tag-prompting mode without
chain-of-thought, not in the same CoT format used by the trained students.
As \citet{kajitsuka-capacitygap-2026} demonstrate, comparisons
between distilled students and teachers without proper pre-distillation baselines
can be misleading; distillation can sometimes degrade baseline performance.

\paragraph{Mechanism ambiguity.}
StudentF's accuracy gains over the teacher are consistent with both genuine
knowledge transfer from hindsight-guided traces and format
alignment~\cite{yang-emperors-2025}. ~\citet{wadhwa-mysteries-2024}
showed that CoT distillation gains can survive shuffled rationales, suggesting
that format signal alone can drive improvement. Our measurements quantify the net
accuracy effect but cannot distinguish the underlying mechanism.

\paragraph{Quantisation.}
Student and StudentF are evaluated at 4-bit \ConstStudentQuant{} while the
Base (tagged) reference runs at full floating-point precision.
This asymmetry is conservative for fine-tuned models: any advantage over
Base is observed despite a quantisation disadvantage.
Cross-format effects between student \ConstStudentQuant{} and teacher
\ConstTeacherQuant{} are unmeasured.

\paragraph{Statistical clustering.}
Tests and CIs cluster by disease (\ConstClusterN{} clusters,
\ConstZebraMapEvalCasesPerDisease{} cases each).
ICC(1) values of 0.17--0.20 empirically confirm that disease-level clustering
is warranted; however, strict exchangeability does not hold---diseases
that are hard for one model tend to be hard for all
(same $\bar{\rho}{=}\StatMeanPairwiseRho{}$ concordance across model pairs),
violating the assumption that clusters are interchangeable draws from a uniform
difficulty distribution. CIs are
therefore conservative approximations rather than exact intervals.

\paragraph{Training randomness.}
A single Student and a single StudentF are reported. Evaluation CIs are
clustered over disease strata; variance across training runs is not quantified.

%==============================================================================
\bibliography{custom}

@misc{hinton-distill-2015,
    title={Distilling the Knowledge in a Neural Network}, 
    author={Geoffrey Hinton and Oriol Vinyals and Jeff Dean},
    year={2015},
    eprint={1503.02531},
    archivePrefix={arXiv},
    primaryClass={stat.ML},
    url={https://arxiv.org/abs/1503.02531}, 
}

@inproceedings{wei-etal-2022-chain,
 author = {Wei, Jason and Wang, Xuezhi and Schuurmans, Dale and Bosma, Maarten and Ichter, Brian and Xia, Fei and Chi, Ed and Le, Quoc V and Zhou, Denny},
 booktitle = {Advances in Neural Information Processing Systems},
 editor = {S. Koyejo and S. Mohamed and A. Agarwal and D. Belgrave and K. Cho and A. Oh},
 pages = {24824--24837},
 publisher = {Curran Associates, Inc.},
 title = {Chain-of-Thought Prompting Elicits Reasoning in Large Language Models},
 url = {https://proceedings.neurips.cc/paper_files/paper/2022/file/9d5609613524ecf4f15af0f7b31abca4-Paper-Conference.pdf},
 volume = {35},
 year = {2022}
}

@inproceedings{ho-etal-2023-large,
    title = "Large Language Models Are Reasoning Teachers",
    author = "Ho, Namgyu  and
      Schmid, Laura  and
      Yun, Se-Young",
    editor = "Rogers, Anna  and
      Boyd-Graber, Jordan  and
      Okazaki, Naoaki",
    booktitle = "Proceedings of the 61st Annual Meeting of the Association for Computational Linguistics (Volume 1: Long Papers)",
    month = jul,
    year = "2023",
    address = "Toronto, Canada",
    publisher = "Association for Computational Linguistics",
    url = "https://aclanthology.org/2023.acl-long.830/",
    doi = "10.18653/v1/2023.acl-long.830",
    pages = "14852--14882"
}

@inproceedings{magister-etal-2023-teaching,
    title = "Teaching Small Language Models to Reason",
    author = "Magister, Lucie Charlotte  and
      Mallinson, Jonathan  and
      Adamek, Jakub  and
      Malmi, Eric  and
      Severyn, Aliaksei",
    editor = "Rogers, Anna  and
      Boyd-Graber, Jordan  and
      Okazaki, Naoaki",
    booktitle = "Proceedings of the 61st Annual Meeting of the Association for Computational Linguistics (Volume 2: Short Papers)",
    month = jul,
    year = "2023",
    address = "Toronto, Canada",
    publisher = "Association for Computational Linguistics",
    url = "https://aclanthology.org/2023.acl-short.151/",
    doi = "10.18653/v1/2023.acl-short.151",
    pages = "1773--1781"
}

@inproceedings{hsieh-etal-2023-distilling,
    title = "Distilling Step-by-Step! Outperforming Larger Language Models with Less Training Data and Smaller Model Sizes",
    author = "Hsieh, Cheng-Yu  and
      Li, Chun-Liang  and
      Yeh, Chih-kuan  and
      Nakhost, Hootan  and
      Fujii, Yasuhisa  and
      Ratner, Alex  and
      Krishna, Ranjay  and
      Lee, Chen-Yu  and
      Pfister, Tomas",
    editor = "Rogers, Anna  and
      Boyd-Graber, Jordan  and
      Okazaki, Naoaki",
    booktitle = "Findings of the Association for Computational Linguistics: ACL 2023",
    month = jul,
    year = "2023",
    address = "Toronto, Canada",
    publisher = "Association for Computational Linguistics",
    url = "https://aclanthology.org/2023.findings-acl.507/",
    doi = "10.18653/v1/2023.findings-acl.507",
    pages = "8003--8017"
}

@inproceedings{wang-scott-2023,
    title = "{SCOTT}: Self-Consistent Chain-of-Thought Distillation",
    author = "Wang, Peifeng  and
      Wang, Zhengyang  and
      Li, Zheng  and
      Gao, Yifan  and
      Yin, Bing  and
      Ren, Xiang",
    editor = "Rogers, Anna  and
      Boyd-Graber, Jordan  and
      Okazaki, Naoaki",
    booktitle = "Proceedings of the 61st Annual Meeting of the Association for Computational Linguistics (Volume 1: Long Papers)",
    month = jul,
    year = "2023",
    address = "Toronto, Canada",
    publisher = "Association for Computational Linguistics",
    url = "https://aclanthology.org/2023.acl-long.304/",
    doi = "10.18653/v1/2023.acl-long.304",
    pages = "5546--5558"
}

@misc{wang-selfconsistency-2022,
      title={Self-Consistency Improves Chain of Thought Reasoning in Language Models}, 
      author={Xuezhi Wang and Jason Wei and Dale Schuurmans and Quoc Le and Ed Chi and Sharan Narang and Aakanksha Chowdhery and Denny Zhou},
      year={2023},
      eprint={2203.11171},
      archivePrefix={arXiv},
      primaryClass={cs.CL},
      url={https://arxiv.org/abs/2203.11171}, 
}

@misc{li-confidence-2024,
      title={Confidence Matters: Revisiting Intrinsic Self-Correction Capabilities of Large Language Models}, 
      author={Loka Li and Zhenhao Chen and Guangyi Chen and Yixuan Zhang and Yusheng Su and Eric Xing and Kun Zhang},
      year={2024},
      eprint={2402.12563},
      archivePrefix={arXiv},
      primaryClass={cs.CL},
      url={https://arxiv.org/abs/2402.12563}, 
}

@misc{shinn-reflexion-2023,
      title={Reflexion: Language Agents with Verbal Reinforcement Learning}, 
      author={Noah Shinn and Federico Cassano and Edward Berman and Ashwin Gopinath and Karthik Narasimhan and Shunyu Yao},
      year={2023},
      eprint={2303.11366},
      archivePrefix={arXiv},
      primaryClass={cs.AI},
      url={https://arxiv.org/abs/2303.11366}, 
}

@inproceedings{zelikman-etal-2022-star,
 author = {Zelikman, Eric and Wu, Yuhuai and Mu, Jesse and Goodman, Noah},
 booktitle = {Advances in Neural Information Processing Systems},
 editor = {S. Koyejo and S. Mohamed and A. Agarwal and D. Belgrave and K. Cho and A. Oh},
 pages = {15476--15488},
 publisher = {Curran Associates, Inc.},
 title = {STaR: Bootstrapping Reasoning With Reasoning},
 url = {https://proceedings.neurips.cc/paper_files/paper/2022/file/639a9a172c044fbb64175b5fad42e9a5-Paper-Conference.pdf},
 volume = {35},
 year = {2022}
}

@inproceedings{ouyang-instructgpt-2022,
 author = {Ouyang, Long and Wu, Jeffrey and Jiang, Xu and Almeida, Diogo and Wainwright, Carroll and Mishkin, Pamela and Zhang, Chong and Agarwal, Sandhini and Slama, Katarina and Ray, Alex and Schulman, John and Hilton, Jacob and Kelton, Fraser and Miller, Luke and Simens, Maddie and Askell, Amanda and Welinder, Peter and Christiano, Paul F and Leike, Jan and Lowe, Ryan},
 booktitle = {Advances in Neural Information Processing Systems},
 editor = {S. Koyejo and S. Mohamed and A. Agarwal and D. Belgrave and K. Cho and A. Oh},
 pages = {27730--27744},
 publisher = {Curran Associates, Inc.},
 title = {Training language models to follow instructions with human feedback},
 url = {https://proceedings.neurips.cc/paper_files/paper/2022/file/b1efde53be364a73914f58805a001731-Paper-Conference.pdf},
 volume = {35},
 year = {2022}
}

@misc{rafailov-dpo-2023,
      title={Direct Preference Optimization: Your Language Model is Secretly a Reward Model}, 
      author={Rafael Rafailov and Archit Sharma and Eric Mitchell and Stefano Ermon and Christopher D. Manning and Chelsea Finn},
      year={2024},
      eprint={2305.18290},
      archivePrefix={arXiv},
      primaryClass={cs.LG},
      url={https://arxiv.org/abs/2305.18290}, 
}

@article{sft-or-rl-2025,
    title={{SFT} or {RL}? An Early Investigation into Training R1-Like Reasoning Large Vision-Language Models},
    author={Hardy Chen and Haoqin Tu and Fali Wang and Hui Liu and Xianfeng Tang and Xinya Du and Yuyin Zhou and Cihang Xie},
    journal={Transactions on Machine Learning Research},
    issn={2835-8856},
    year={2025},
    url={https://openreview.net/forum?id=wZI5qkQeDF},
    note={}
}

@misc{turpin-unfaithful-2023,
      title={Language Models Don't Always Say What They Think: Unfaithful Explanations in Chain-of-Thought Prompting}, 
      author={Miles Turpin and Julian Michael and Ethan Perez and Samuel R. Bowman},
      year={2023},
      eprint={2305.04388},
      archivePrefix={arXiv},
      primaryClass={cs.CL},
      url={https://arxiv.org/abs/2305.04388}, 
}

@misc{dettmers-qlora-2023,
      title={QLoRA: Efficient Finetuning of Quantized LLMs}, 
      author={Tim Dettmers and Artidoro Pagnoni and Ari Holtzman and Luke Zettlemoyer},
      year={2023},
      eprint={2305.14314},
      archivePrefix={arXiv},
      primaryClass={cs.LG},
      url={https://arxiv.org/abs/2305.14314},
}

@inproceedings{zhou-lima-2023,
 author = {Zhou, Chunting and Liu, Pengfei and Xu, Puxin and Iyer, Srinivasan and Sun, Jiao and Mao, Yuning and Ma, Xuezhe and Efrat, Avia and Yu, Ping and Yu, Lili and Zhang, Susan and Ghosh, Gargi and Lewis, Mike and Zettlemoyer, Luke and Levy, Omer},
 booktitle = {Advances in Neural Information Processing Systems},
 editor = {A. Oh and T. Naumann and A. Globerson and K. Saenko and M. Hardt and S. Levine},
 pages = {55006--55021},
 publisher = {Curran Associates, Inc.},
 title = {LIMA: Less Is More for Alignment},
 url = {https://proceedings.neurips.cc/paper_files/paper/2023/file/ac662d74829e4407ce1d126477f4a03a-Paper-Conference.pdf},
 volume = {36},
 year = {2023}
}

@Article{deepseek-r1-2025,
author={Guo, Daya
and Yang, Dejian
and Zhang, Haowei
and Song, Junxiao
and Wang, Peiyi
and Zhu, Qihao
and Xu, Runxin
and Zhang, Ruoyu
and Ma, Shirong
and Bi, Xiao
and Zhang, Xiaokang
and Yu, Xingkai
and Wu, Yu
and Wu, Z. F.
and Gou, Zhibin
and Shao, Zhihong
and Li, Zhuoshu
and Gao, Ziyi
and Liu, Aixin
and Xue, Bing
and Wang, Bingxuan
and Wu, Bochao
and Feng, Bei
and Lu, Chengda
and Zhao, Chenggang
and Deng, Chengqi
and Ruan, Chong
and Dai, Damai
and Chen, Deli
and Ji, Dongjie
and Li, Erhang
and Lin, Fangyun
and Dai, Fucong
and Luo, Fuli
and Hao, Guangbo
and Chen, Guanting
and Li, Guowei
and Zhang, H.
and Xu, Hanwei
and Ding, Honghui
and Gao, Huazuo
and Qu, Hui
and Li, Hui
and Guo, Jianzhong
and Li, Jiashi
and Chen, Jingchang
and Yuan, Jingyang
and Tu, Jinhao
and Qiu, Junjie
and Li, Junlong
and Cai, J. L.
and Ni, Jiaqi
and Liang, Jian
and Chen, Jin
and Dong, Kai
and Hu, Kai
and You, Kaichao
and Gao, Kaige
and Guan, Kang
and Huang, Kexin
and Yu, Kuai
and Wang, Lean
and Zhang, Lecong
and Zhao, Liang
and Wang, Litong
and Zhang, Liyue
and Xu, Lei
and Xia, Leyi
and Zhang, Mingchuan
and Zhang, Minghua
and Tang, Minghui
and Zhou, Mingxu
and Li, Meng
and Wang, Miaojun
and Li, Mingming
and Tian, Ning
and Huang, Panpan
and Zhang, Peng
and Wang, Qiancheng
and Chen, Qinyu
and Du, Qiushi
and Ge, Ruiqi
and Zhang, Ruisong
and Pan, Ruizhe
and Wang, Runji
and Chen, R. J.
and Jin, R. L.
and Chen, Ruyi
and Lu, Shanghao
and Zhou, Shangyan
and Chen, Shanhuang
and Ye, Shengfeng
and Wang, Shiyu
and Yu, Shuiping
and Zhou, Shunfeng
and Pan, Shuting
and Li, S. S.
and Zhou, Shuang
and Wu, Shaoqing
and Yun, Tao
and Pei, Tian
and Sun, Tianyu
and Wang, T.
and Zeng, Wangding
and Liu, Wen
and Liang, Wenfeng
and Gao, Wenjun
and Yu, Wenqin
and Zhang, Wentao
and Xiao, W. L.
and An, Wei
and Liu, Xiaodong
and Wang, Xiaohan
and Chen, Xiaokang
and Nie, Xiaotao
and Cheng, Xin
and Liu, Xin
and Xie, Xin
and Liu, Xingchao
and Yang, Xinyu
and Li, Xinyuan
and Su, Xuecheng
and Lin, Xuheng
and Li, X. Q.
and Jin, Xiangyue
and Shen, Xiaojin
and Chen, Xiaosha
and Sun, Xiaowen
and Wang, Xiaoxiang
and Song, Xinnan
and Zhou, Xinyi
and Wang, Xianzu
and Shan, Xinxia
and Li, Y. K.
and Wang, Y. Q.
and Wei, Y. X.
and Zhang, Yang
and Xu, Yanhong
and Li, Yao
and Zhao, Yao
and Sun, Yaofeng
and Wang, Yaohui
and Yu, Yi
and Zhang, Yichao
and Shi, Yifan
and Xiong, Yiliang
and He, Ying
and Piao, Yishi
and Wang, Yisong
and Tan, Yixuan
and Ma, Yiyang
and Liu, Yiyuan
and Guo, Yongqiang
and Ou, Yuan
and Wang, Yuduan
and Gong, Yue
and Zou, Yuheng
and He, Yujia
and Xiong, Yunfan
and Luo, Yuxiang
and You, Yuxiang
and Liu, Yuxuan
and Zhou, Yuyang
and Zhu, Y. X.
and Huang, Yanping
and Li, Yaohui
and Zheng, Yi
and Zhu, Yuchen
and Ma, Yunxian
and Tang, Ying
and Zha, Yukun
and Yan, Yuting
and Ren, Z. Z.
and Ren, Zehui
and Sha, Zhangli
and Fu, Zhe
and Xu, Zhean
and Xie, Zhenda
and Zhang, Zhengyan
and Hao, Zhewen
and Ma, Zhicheng
and Yan, Zhigang
and Wu, Zhiyu
and Gu, Zihui
and Zhu, Zijia
and Liu, Zijun
and Li, Zilin
and Xie, Ziwei
and Song, Ziyang
and Pan, Zizheng
and Huang, Zhen
and Xu, Zhipeng
and Zhang, Zhongyu
and Zhang, Zhen},
title={DeepSeek-R1 incentivizes reasoning in LLMs through reinforcement learning},
journal={Nature},
year={2025},
month={Sep},
day={01},
volume={645},
number={8081},
pages={633-638},
issn={1476-4687},
doi={10.1038/s41586-025-09422-z},
url={https://doi.org/10.1038/s41586-025-09422-z}
}

@misc{shao-deepseekmath-2024,
      title={DeepSeekMath: Pushing the Limits of Mathematical Reasoning in Open Language Models}, 
      author={Zhihong Shao and Peiyi Wang and Qihao Zhu and Runxin Xu and Junxiao Song and Xiao Bi and Haowei Zhang and Mingchuan Zhang and Y. K. Li and Y. Wu and Daya Guo},
      year={2024},
      eprint={2402.03300},
      archivePrefix={arXiv},
      primaryClass={cs.CL},
      url={https://arxiv.org/abs/2402.03300}, 
}

@misc{qwen25-2024,
      title={Qwen2.5 Technical Report}, 
      author={Qwen, Team},
      year={2025},
      eprint={2412.15115},
      archivePrefix={arXiv},
      primaryClass={cs.CL},
      url={https://arxiv.org/abs/2412.15115}, 
}

@article{singhal-medpalm-2023,
author={Singhal, Karan
and Azizi, Shekoofeh
and Tu, Tao
and Mahdavi, S. Sara
and Wei, Jason
and Chung, Hyung Won
and Scales, Nathan
and Tanwani, Ajay
and Cole-Lewis, Heather
and Pfohl, Stephen
and Payne, Perry
and Seneviratne, Martin
and Gamble, Paul
and Kelly, Chris
and Babiker, Abubakr
and Sch{\"a}rli, Nathanael
and Chowdhery, Aakanksha
and Mansfield, Philip
and Demner-Fushman, Dina
and Ag{\"u}era y Arcas, Blaise
and Webster, Dale
and Corrado, Greg S.
and Matias, Yossi
and Chou, Katherine
and Gottweis, Juraj
and Tomasev, Nenad
and Liu, Yun
and Rajkomar, Alvin
and Barral, Joelle
and Semturs, Christopher
and Karthikesalingam, Alan
and Natarajan, Vivek},
title={Large language models encode clinical knowledge},
journal={Nature},
year={2023},
month={Aug},
day={01},
volume={620},
number={7972},
pages={172-180},
issn={1476-4687},
doi={10.1038/s41586-023-06291-2},
url={https://doi.org/10.1038/s41586-023-06291-2}
}

@article{ao-llm-raredisease-2025,
  title={Comparative analysis of large language models on rare disease identification},
  author={Ao, Guangyu and Chen, Min and Li, Jing and Nie, Huibing and Zhang, Lei and Chen, Zejun},
  journal={Orphanet Journal of Rare Diseases},
  volume={20},
  number={1},
  pages={150},
  year={2025},
  publisher={Springer},
  doi={10.1186/s13023-025-03656-w},
}

@article{young-llm-rare-2025,
    author = {Young, Cameron C. and Enichen, Ellie and Rivera, Christian and Auger, Corinne A. and Grant, Nathan and Rao, Arya and Succi, Marc D.},
    title = {Diagnostic Accuracy of a Custom Large Language Model on Rare Pediatric Disease Case Reports},
    journal = {American Journal of Medical Genetics Part A},
    volume = {197},
    number = {2},
    pages = {e63878},
    doi = {https://doi.org/10.1002/ajmg.a.63878},
    url = {https://onlinelibrary.wiley.com/doi/abs/10.1002/ajmg.a.63878},
    eprint = {https://onlinelibrary.wiley.com/doi/pdf/10.1002/ajmg.a.63878},
    year = {2025}
}

@article{benchmarking-llm-rare-2025,
    author={Reese, Justin T. and Chimirri, Leonardo and Bridges, Yasemin and Danis, Daniel and Caufield, J. Harry and Gargano, Michael A. and Kroll, Carlo and Schmeder, Andrew and Liu, Fengchen and Wissink, Kyran and McMurry, Julie A. and Graefe, Adam S. L. and Niyonkuru, Enock and Korn, Daniel R. and Casiraghi, Elena and Valentini, Giorgio and Jacobsen, Julius O. B. and Haendel, Melissa and Smedley, Damian and Mungall, Christopher J. and Robinson, Peter N.},
    title={Systematic benchmarking demonstrates large language models have not reached the diagnostic accuracy of traditional rare-disease decision support tools},
    journal={European Journal of Human Genetics},
    year={2026},
    month={Apr},
    day={01},
    volume={34},
    number={4},
    pages={498-504},
    issn={1476-5438},
    doi={10.1038/s41431-026-02054-5},
    url={https://doi.org/10.1038/s41431-026-02054-5}
}

@article{segura-raredis-2022,
author={Segura-Bedmar, Isabel
and Camino-Perdones, David
and Guerrero-Aspizua, Sara},
title={Exploring deep learning methods for recognizing rare diseases and their clinical manifestations from texts},
journal={BMC Bioinformatics},
year={2022},
month={Jul},
day={06},
volume={23},
number={1},
pages={263},
issn={1471-2105},
doi={10.1186/s12859-022-04810-y},
url={https://doi.org/10.1186/s12859-022-04810-y}
}

@article{zebramap2026,
AUTHOR = {Islam, Md. Sanzidul and Jamal, Amani and Alkhathlan, Ali},
TITLE = {ZebraMap: A Multimodal Rare Disease Knowledge Map with Automated Data Aggregation \& LLM-Enriched Information Extraction Pipeline},
JOURNAL = {Diagnostics},
VOLUME = {16},
YEAR = {2026},
NUMBER = {1},
ARTICLE-NUMBER = {107},
URL = {https://www.mdpi.com/2075-4418/16/1/107},
PubMedID = {41515602},
ISSN = {2075-4418},
DOI = {10.3390/diagnostics16010107}
}

@Article{s26113582,
  AUTHOR = {Islam, Md Sanzidul and Jamal, Amani and Alkhathlan, Ali},
  TITLE = {Rare-Disease Diagnosis on the ZebraMap Multimodal Case Report Dataset: A Hybrid Pipeline with Grounded Explainability},
  JOURNAL = {Sensors},
  VOLUME = {26},
  YEAR = {2026},
  NUMBER = {11},
  ARTICLE-NUMBER = {3582},
  URL = {https://www.mdpi.com/1424-8220/26/11/3582},
  ISSN = {1424-8220},
  DOI = {10.3390/s26113582}
}

@article{wakap-prevalence-2020,
    author={Nguengang Wakap, St{\'e}phanie
    and Lambert, Deborah M.
    and Olry, Annie
    and Rodwell, Charlotte
    and Gueydan, Charlotte
    and Lanneau, Val{\'e}rie
    and Murphy, Daniel
    and Le Cam, Yann
    and Rath, Ana},
    title={Estimating cumulative point prevalence of rare diseases: analysis of the Orphanet database},
    journal={European Journal of Human Genetics},
    year={2020},
    month={Feb},
    day={01},
    volume={28},
    number={2},
    pages={165-173},
    issn={1476-5438},
    doi={10.1038/s41431-019-0508-0},
    url={https://doi.org/10.1038/s41431-019-0508-0}
}

@article{bruckner-epidemiology-2021,
author = {Bruckner-Tuderman, L.},
title = {Epidemiology of rare diseases is important},
journal = {Journal of the European Academy of Dermatology and Venereology},
volume = {35},
number = {4},
pages = {783-784},
doi = {https://doi.org/10.1111/jdv.17165},
url = {https://onlinelibrary.wiley.com/doi/abs/10.1111/jdv.17165},
eprint = {https://onlinelibrary.wiley.com/doi/pdf/10.1111/jdv.17165},
year = {2021}
}

@misc{xu-kd-survey-2024,
      title={A Survey on Knowledge Distillation of Large Language Models}, 
      author={Xiaohan Xu and Ming Li and Chongyang Tao and Tao Shen and Reynold Cheng and Jinyang Li and Can Xu and Dacheng Tao and Tianyi Zhou},
      year={2024},
      eprint={2402.13116},
      archivePrefix={arXiv},
      primaryClass={cs.CL},
      url={https://arxiv.org/abs/2402.13116}, 
}

@inproceedings{zhuang-efficient-medical-2025,
    author    = {Zhuang, Xinlin and Tang, Feilong and Yang, Haolin and Liu, Xiwei and Hu, Ming and Li, Huifa and Xue, Haochen and He, Junjun and Ge, Zongyuan and Li, Yichen and Qian, Ying and Razzak, Imran},
    title     = {Towards Efficient Medical Reasoning with Minimal Fine-Tuning Data},
    booktitle = {Proceedings of the IEEE/CVF Conference on Computer Vision and Pattern Recognition (CVPR)},
    month     = {June},
    year      = {2026},
    pages     = {20223-20232},
    url = {https://openaccess.thecvf.com/content/CVPR2026/papers/Zhuang_Towards_Efficient_Medical_Reasoning_with_Minimal_Fine-Tuning_Data_CVPR_2026_paper.pdf},
}

@misc{wu-medreason-2025,
      title={MedReason: Eliciting Factual Medical Reasoning Steps in LLMs via Knowledge Graphs}, 
      author={Juncheng Wu and Wenlong Deng and Xingxuan Li and Sheng Liu and Taomian Mi and Yifan Peng and Ziyang Xu and Yi Liu and Hyunjin Cho and Chang-In Choi and Yihan Cao and Hui Ren and Xiang Li and Xiaoxiao Li and Yuyin Zhou},
      year={2025},
      eprint={2504.00993},
      archivePrefix={arXiv},
      primaryClass={cs.CL},
      url={https://arxiv.org/abs/2504.00993}, 
}

@inproceedings{green-leaky-2025,
    title = "Leaky Thoughts: Large Reasoning Models Are Not Private Thinkers",
    author = "Green, Tommaso  and
      Gubri, Martin  and
      Puerto, Haritz  and
      Yun, Sangdoo  and
      Oh, Seong Joon",
    editor = "Christodoulopoulos, Christos  and
      Chakraborty, Tanmoy  and
      Rose, Carolyn  and
      Peng, Violet",
    booktitle = "Proceedings of the 2025 Conference on Empirical Methods in Natural Language Processing",
    month = nov,
    year = "2025",
    address = "Suzhou, China",
    publisher = "Association for Computational Linguistics",
    url = "https://aclanthology.org/2025.emnlp-main.1347/",
    doi = "10.18653/v1/2025.emnlp-main.1347",
    pages = "26507--26529",
    ISBN = "979-8-89176-332-6"
}

@misc{jiang-betteracc-2026,
      title={Better Accuracies, Worse Reasoning: A Step-Level Audit of Medical Chain-of-Thought Distillation}, 
      author={Zhaoyang Jiang and Xuanqi Peng and Fei Teng and Zhizhong Fu and Yunsoo Kim and Jiacong Mi and Zicheng Li and Honghan Wu},
      year={2026},
      eprint={2605.28301},
      archivePrefix={arXiv},
      primaryClass={cs.AI},
      url={https://arxiv.org/abs/2605.28301}, 
}

@misc{kajitsuka-capacitygap-2026,
      title={Revisiting the Capacity Gap in Chain-of-Thought Distillation from a Practical Perspective}, 
      author={Tokio Kajitsuka and Ukyo Honda and Sho Takase},
      year={2026},
      eprint={2604.08880},
      archivePrefix={arXiv},
      primaryClass={cs.LG},
      url={https://arxiv.org/abs/2604.08880}, 
}

@inproceedings{yang-emperors-2025,
    title = "The Emperor{'}s New Reasoning: Format Imitation Overshadows Genuine Mathematical Understanding in {SFT}",
    author = "Yang, Linyao  and
      Huang, Jian-Tao  and
      Lu, Yafei  and
      Li, Zhenhui Jessie  and
      Xue, Guirong",
    editor = "Christodoulopoulos, Christos  and
      Chakraborty, Tanmoy  and
      Rose, Carolyn  and
      Peng, Violet",
    booktitle = "Proceedings of the 2025 Conference on Empirical Methods in Natural Language Processing",
    month = nov,
    year = "2025",
    address = "Suzhou, China",
    publisher = "Association for Computational Linguistics",
    url = "https://aclanthology.org/2025.emnlp-main.1067/",
    doi = "10.18653/v1/2025.emnlp-main.1067",
    pages = "21087--21100",
    ISBN = "979-8-89176-332-6"
}

@inproceedings{chen-keyfactors-2025,
    title = "Unveiling the Key Factors for Distilling Chain-of-Thought Reasoning",
    author = "Chen, Xinghao  and
      Sun, Zhijing  and
      Guo, Wenjin  and
      Zhang, Miaoran  and
      Chen, Yanjun  and
      Sun, Yirong  and
      Su, Hui  and
      Pan, Yijie  and
      Klakow, Dietrich  and
      Li, Wenjie  and
      Shen, Xiaoyu",
    editor = "Che, Wanxiang  and
      Nabende, Joyce  and
      Shutova, Ekaterina  and
      Pilehvar, Mohammad Taher",
    booktitle = "Findings of the Association for Computational Linguistics: ACL 2025",
    month = jul,
    year = "2025",
    address = "Vienna, Austria",
    publisher = "Association for Computational Linguistics",
    url = "https://aclanthology.org/2025.findings-acl.782/",
    doi = "10.18653/v1/2025.findings-acl.782",
    pages = "15094--15119",
    ISBN = "979-8-89176-256-5"
}

@inproceedings{wadhwa-mysteries-2024,
    title = "Investigating Mysteries of {C}o{T}-Augmented Distillation",
    author = "Wadhwa, Somin  and
      Amir, Silvio  and
      Wallace, Byron C",
    editor = "Al-Onaizan, Yaser  and
      Bansal, Mohit  and
      Chen, Yun-Nung",
    booktitle = "Proceedings of the 2024 Conference on Empirical Methods in Natural Language Processing",
    month = nov,
    year = "2024",
    address = "Miami, Florida, USA",
    publisher = "Association for Computational Linguistics",
    url = "https://aclanthology.org/2024.emnlp-main.349/",
    doi = "10.18653/v1/2024.emnlp-main.349",
    pages = "6071--6086"
}

@misc{dccot-2025,
title={The Quest for Efficient Reasoning: A Data-Centric Benchmark to CoT Distillation},
author={Ruichen Zhang and Rana Shahroz and Zhen Tan and Dawei Li and Song Wang and Tianlong Chen},
booktitle={The Fourteenth International Conference on Learning Representations},
year={2026},
url={https://openreview.net/forum?id=Dud8FtScW7}

}

@article{rareaena-lancet-2025,
    author={Chen, Haichao and Zhao, Zhengyun and Zhou, Songchi and Hu, Shikai and Wang, Jinyuan and Jin, Ye and Jin, Xianghong and Tham, Yih Chung and Wang, Xiaofei and Ma, Weizhi and Wu, Honghan and Sheng, Bin and Zhang, Shuyang and Yu, Sheng and Wong, Tien Yin},
    title={RareArena: a comprehensive benchmark dataset unveiling the potential of large language models in rare disease diagnosis},
    journal={The Lancet Digital Health},
    year={2026},
    month={Feb},
    day={01},
    publisher={Elsevier},
    volume={8},
    number={2},
    issn={2589-7500},
    doi={10.1016/j.landig.2025.100953},
    url={https://doi.org/10.1016/j.landig.2025.100953}
}

@Misc{medr1-2025,
  title = {Med-R1: Encourage Medical LLM to engage in deep thinking similar to DeepSeek-R1},
  author = {Rongsheng Wang},
  howpublished = {\url{https://github.com/WangRongsheng/Med-R1}},
  year = {2025}
}

@misc{wu-cotfails-2025a,
      title={Why Chain of Thought Fails in Clinical Text Understanding}, 
      author={Jiageng Wu and Kevin Xie and Bowen Gu and Nils Krüger and Kueiyu Joshua Lin and Jie Yang},
      year={2025},
      eprint={2509.21933},
      archivePrefix={arXiv},
      primaryClass={cs.CL},
      url={https://arxiv.org/abs/2509.21933}, 
}

@inproceedings{yang-srft-2025,
title={{SRFT}: A Single-Stage Method with Supervised and Reinforcement Fine-Tuning for Reasoning},
author={Yuqian Fu and Tinghong Chen and Jiajun Chai and Xihuai Wang and Songjun Tu and Guojun Yin and Wei Lin and Qichao Zhang and Yuanheng Zhu and Dongbin Zhao},
booktitle={The Fourteenth International Conference on Learning Representations},
year={2026},
url={https://openreview.net/forum?id=n6E0r6kQWQ}
}

@inproceedings{kstar-aaai-2026,
  author    = {Li, Guozheng and Zhang, Xinyu},
  title     = {{K-STaR}: Knowledge-Aware Self-Taught Reasoner},
  booktitle = {Proceedings of the AAAI Conference on Artificial Intelligence},
  volume    = {40},
  year      = {2026},
  pages     = {31573--31581},
  publisher = {AAAI Press},
  doi       = {10.1609/aaai.v40i37.40423},
  url       = {https://ojs.aaai.org/index.php/AAAI/article/view/40423},
}

%==============================================================================
\FloatBarrier
\appendix

\section{Error Taxonomy}
\label{app:errors}

\paragraph{Category definitions (priority cascade).}
Each response is assigned to exactly one category, evaluated in the order
listed below. The first matching criterion determines the label.

\begin{enumerate}[leftmargin=*,label=(\arabic*),itemsep=1pt,topsep=2pt]
\item \textbf{Correct (M3):} The \texttt{<diagnosis>} tag content matches
      the ground-truth disease via bidirectional synonym match (the primary
      Accuracy metric).
\item \textbf{Format failure — recovered:} No valid tag, but the correct
      disease appears in post-think prose (captured by Accuracy, upper bound).
\item \textbf{GT-poisoned:} Response contains a ``ground truth is X''
      phrase with the wrong disease, which is then copied into the
      \texttt{<diagnosis>} tag. Student-exclusive artifact.
\item \textbf{Revised away:} The correct disease appears in the think
      chain (M1-correct) but the model self-revised away from it before
      committing to a wrong \texttt{<diagnosis>} tag.
\item \textbf{Near-miss:} The predicted disease is wrong by M3, but its
      embedding has cosine similarity $> 0.7$ with the true disease
      embedding (semantically very close).
\item \textbf{Weak related:} Cosine similarity $0.5 < \cos \leq 0.7$ to
      the true disease (semantically related but clearly distinct).
\item \textbf{Wrong — same ICD-10 category:} Prediction and ground truth
      are in the same ICD-10 chapter but cosine similarity $\leq 0.5$.
\item \textbf{Wrong — different ICD-10 category:} Prediction and ground
      truth are in different ICD-10 chapters; both diseases have ICD-10 codes.
\item \textbf{Both uncategorized:} Neither the predicted nor the true
      disease has an assigned ICD-10 code. Occurs for very rare diseases
      not yet in the ICD-10 registry.
\item \textbf{Prediction uncategorized:} The predicted disease has no
      ICD-10 code, but the true disease does.
\item \textbf{True uncategorized:} The true disease has no ICD-10 code,
      but the predicted disease does.
\item \textbf{Format failure — unrecoverable:} No valid tag and no
      recoverable answer in post-think prose.
\end{enumerate}

Categories (9)--(11) collectively cover all wrong predictions where at
least one of the two diseases (predicted or true) lacks an ICD-10 code,
reflecting the prevalence of extremely rare diseases in ZebraMap that are
not yet assigned ICD-10 codes. These constitute a substantial fraction of
wrong-answer cases for all models.

% Error taxonomy — stacked horizontal bar chart (pipeline A.6 categories)
% Priority cascade: Correct → Format(recovered) → GT-poisoned → Near-miss
%   → Weak related → Wrong(same ICD-10) → Wrong(diff ICD-10)
%   → Revised away → ICD-10 gap (3 sub-types) → Format(wrong)

\definecolor{colCorrect}{RGB}{75,160,100}        % green
\definecolor{colFmtKnown}{RGB}{160,210,160}      % light green
\definecolor{colGTPois}{RGB}{199,52,63}          % red
\definecolor{colRevised}{RGB}{74,130,185}       % blue
\definecolor{colWeakRel}{RGB}{140,190,215}       % light blue
\definecolor{colWrongSame}{RGB}{237,150,60}      % orange
\definecolor{colWrongDiff}{RGB}{190,100,30}      % dark orange
\definecolor{colNearMiss}{RGB}{180,120,200}       % purple
\definecolor{colOtherOther}{RGB}{220,200,230}    % lavender
\definecolor{colOtherKnown}{RGB}{190,170,210}    % medium lavender
\definecolor{colKnownOther}{RGB}{150,130,175}    % dark lavender
\definecolor{colFmtUnk}{RGB}{180,180,180}        % grey

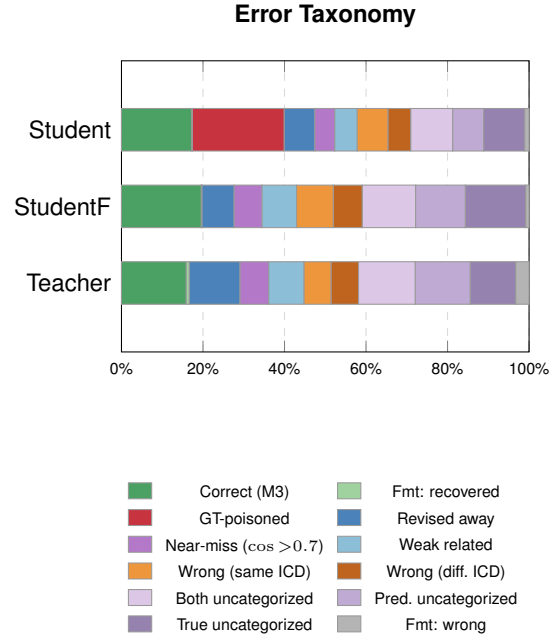
\begin{figure}[htb]
\centering
\begin{tikzpicture}
\begin{axis}[
    xbar stacked,
    width=0.7\columnwidth,
    height=0.50\columnwidth,
    xmin=0, xmax=1.0,
    xtick={0,0.2,0.4,0.6,0.8,1.0},
    xticklabels={0\%,20\%,40\%,60\%,80\%,100\%},
    xticklabel style={font=\tiny\sffamily},
    ytick={0,1,2},
    yticklabels={Teacher, StudentF, Student},
    yticklabel style={font=\small\sffamily, align=right},
    bar width=16pt,
    scale only axis,
    enlarge y limits=0.45,
    xmajorgrids=true,
    grid style={dashed,gray!30},
    title={Error Taxonomy},
    title style={font=\small\bfseries\sffamily, yshift=3pt},
    legend style={
        at={(0.5,-0.42)},
        anchor=north,
        legend columns=2,
        draw=none,
        font=\tiny\sffamily,
        column sep=3pt,
    },
]

% 1. Correct (M3)
\addplot[fill=colCorrect,draw=black!40] coordinates
  {(\StatErrTaxCorrectT,0) (\StatErrTaxCorrectSF,1) (\StatErrTaxCorrectS,2)};

% 2. Format failure — correct answer recovered
\addplot[fill=colFmtKnown,draw=black!40] coordinates
  {(\StatErrTaxFormatKnownT,0) (\StatErrTaxFormatKnownSF,1) (\StatErrTaxFormatKnownS,2)};

% 3. GT-poisoned (Student-exclusive)
\addplot[fill=colGTPois,draw=black!40] coordinates
  {(\StatErrTaxGtPoisonedT,0) (\StatErrTaxGtPoisonedSF,1) (\StatErrTaxGtPoisonedS,2)};

% 4. Revised away (correct appeared in think chain, self-revised)
\addplot[fill=colRevised,draw=black!40] coordinates
  {(\StatErrTaxUnexplainedMoneCorrectT,0)
   (\StatErrTaxUnexplainedMoneCorrectSF,1)
   (\StatErrTaxUnexplainedMoneCorrectS,2)};

% 5. Near-miss (cosine > 0.7)
\addplot[fill=colNearMiss,draw=black!40] coordinates
  {(\StatErrTaxNearMissT,0) (\StatErrTaxNearMissSF,1) (\StatErrTaxNearMissS,2)};

% 6. Weak related (0.5 < cos ≤ 0.7)
\addplot[fill=colWeakRel,draw=black!40] coordinates
  {(\StatErrTaxWeakRelatedT,0) (\StatErrTaxWeakRelatedSF,1) (\StatErrTaxWeakRelatedS,2)};

% 7. Wrong — same ICD-10 category
\addplot[fill=colWrongSame,draw=black!40] coordinates
  {(\StatErrTaxWrongSameCatT,0) (\StatErrTaxWrongSameCatSF,1) (\StatErrTaxWrongSameCatS,2)};

% 8. Wrong — different ICD-10 category
\addplot[fill=colWrongDiff,draw=black!40] coordinates
  {(\StatErrTaxWrongCatT,0) (\StatErrTaxWrongCatSF,1) (\StatErrTaxWrongCatS,2)};

% 9. Both uncategorized: neither predicted nor true disease has ICD-10 code
\addplot[fill=colOtherOther,draw=black!40] coordinates
  {(\StatErrTaxOtherOtherT,0) (\StatErrTaxOtherOtherSF,1) (\StatErrTaxOtherOtherS,2)};

% 10. Prediction uncategorized: predicted disease has no ICD-10 code; true does
\addplot[fill=colOtherKnown,draw=black!40] coordinates
  {(\StatErrTaxOtherFromKnownT,0) (\StatErrTaxOtherFromKnownSF,1) (\StatErrTaxOtherFromKnownS,2)};

% 11. True uncategorized: true disease has no ICD-10 code; predicted does
\addplot[fill=colKnownOther,draw=black!40] coordinates
  {(\StatErrTaxKnownFromOtherT,0) (\StatErrTaxKnownFromOtherSF,1) (\StatErrTaxKnownFromOtherS,2)};

% 12. Format failure — wrong / unrecoverable
\addplot[fill=colFmtUnk,draw=black!40] coordinates
  {(\StatErrTaxFormatUnknownT,0) (\StatErrTaxFormatUnknownSF,1) (\StatErrTaxFormatUnknownS,2)};

\legend{%
  Correct (M3),
  Fmt: recovered,
  GT-poisoned,
  Revised away,
  Near-miss ($\cos{>}0.7$),
  Weak related,
  Wrong (same ICD),
  Wrong (diff.\ ICD),
  Both uncategorized,
  Pred.\ uncategorized,
  True uncategorized,
  Fmt: wrong%
}

\end{axis}
\end{tikzpicture}
\caption{Error taxonomy following the pipeline A.6 priority cascade (see
Appendix~\ref{app:errors} for category definitions). GT-poisoned is
exclusive to Student ($\fpeval{round(\StatErrTaxGtPoisonedS*100,1)}$\%).
ICD-10 uncategorized buckets (lavender shades): \emph{Both uncategorized}
= neither predicted nor true disease has an ICD-10 code;
\emph{Pred.\ uncategorized} = predicted disease lacks an ICD-10 code but
true disease has one; \emph{True uncategorized} = true disease lacks an
ICD-10 code but predicted disease has one. These cover rare diseases not
yet assigned ICD-10 codes and constitute a large share of wrong-answer cases.}
\label{fig:appendix}
\end{figure}

\section{Bidirectional Match Parameter Sweep}
\label{app:bidir}

\begin{figure}[htb]
\centering
\begin{tikzpicture}[trim axis left, trim axis right]
\begin{axis}[
    width=0.7\columnwidth,
    height=0.54\columnwidth,
    xlabel={Min-char parameter (mc)},
    ylabel={Accuracy (M3)},
    xlabel style={font=\small\sffamily},
    ylabel style={font=\small\sffamily},
    xmin=-0.5, xmax=10.5,
    ymin=0.10, ymax=0.35,
    xtick={0,1,2,3,4,5,6,7,8,9,10},
    xticklabel style={font=\tiny\sffamily},
    yticklabel style={font=\tiny\sffamily},
    ymajorgrids=true, xmajorgrids=true,
    grid style={dashed,gray!30},
    legend style={at={(0.5,-0.26)}, anchor=north, legend columns=3,
        draw=none, font=\small\sffamily, column sep=6pt},
    title={Bidirectional Match Parameter Sweep},
    title style={font=\small\bfseries\sffamily, yshift=3pt},
    scale only axis,
    clip=false,
]

% StudentF
\addplot[color=teal_color, thick, mark=*, mark size=1.5pt] coordinates {
    (0,\StatSweepSFmcZero)
    (1,\StatSweepSFmcOne)
    (2,\StatSweepSFmcTwo)
    (3,\StatSweepSFmcThree)
    (4,\StatSweepSFmcFour)
    (5,\StatSweepSFmcFive)
    (6,\StatSweepSFmcSix)
    (7,\StatSweepSFmcSeven)
    (8,\StatSweepSFmcEight)
    (9,\StatSweepSFmcNine)
    (10,\StatSweepSFmcTen)
};

% Teacher
\addplot[color=red_color, thick, mark=square*, mark size=1.5pt] coordinates {
    (0,\StatSweepTmcZero)
    (1,\StatSweepTmcOne)
    (2,\StatSweepTmcTwo)
    (3,\StatSweepTmcThree)
    (4,\StatSweepTmcFour)
    (5,\StatSweepTmcFive)
    (6,\StatSweepTmcSix)
    (7,\StatSweepTmcSeven)
    (8,\StatSweepTmcEight)
    (9,\StatSweepTmcNine)
    (10,\StatSweepTmcTen)
};

% Student
\addplot[color=blue_color, thick, dashed, mark=triangle*, mark size=1.5pt] coordinates {
    (0,\StatSweepSmcZero)
    (1,\StatSweepSmcOne)
    (2,\StatSweepSmcTwo)
    (3,\StatSweepSmcThree)
    (4,\StatSweepSmcFour)
    (5,\StatSweepSmcFive)
    (6,\StatSweepSmcSix)
    (7,\StatSweepSmcSeven)
    (8,\StatSweepSmcEight)
    (9,\StatSweepSmcNine)
    (10,\StatSweepSmcTen)
};

\legend{StudentF, Teacher, Student}

% Default (mc=None) annotation — uses exact/synonym mode values
\node[align=left, font=\tiny\sffamily, draw=gray!50, fill=white,
      rounded corners=2pt, inner sep=3pt]
    at (axis cs:6.2,0.325)
    {Default (mc\,=\,None):\;SF\,=\,\StatSweepSFExact{},\;T\,=\,\StatSweepTExact{}\\
     S\,=\,\StatSweepSExact{};\;SF$-$T gap\,=\,$\StatSweepSFminusTExact{}$};

\end{axis}
\end{tikzpicture}
\caption{All three models' Accuracy (M3) across bidirectional match strictness
levels (mc\,=\,0--10). The StudentF advantage over Teacher is directionally
consistent at every setting; the SF$-$T gap grows from
$\StatSweepSFminusTmcZero{}$ at mc\,=\,0 to
$\StatSweepSFminusTmcSix{}$--$\StatSweepSFminusTmcFive{}$ at mc\,=\,5--7,
peaking at $\StatSweepSFminusTExact{}$ at the exact/synonym default
(mc\,=\,None, dashed reference line in annotation). At low mc\,=\,0--4,
any-length substring matches inflate all three models' scores equally,
compressing the detectable gap. The default setting is used throughout
the main paper.}
\label{fig:sweep}
\end{figure}
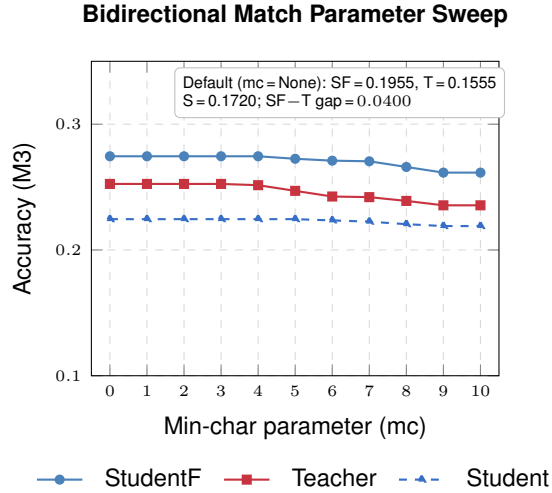

The primary Accuracy metric uses a bidirectional synonym-list match with no minimum
character constraint (\emph{mc}\,=\,None; default). Figure~\ref{fig:sweep} shows
StudentF and Teacher Accuracy across mc\,=\,0--10. The StudentF advantage is
directionally consistent at every setting (SF\,$>$\,T across all 22 tested values
including mc\,=\,None). The SF$-$T gap grows as substring leniency increases:
from $\StatSweepSFminusTmcZero{}$ at mc\,=\,0--3 to
$\StatSweepSFminusTmcFive{}$--$\StatSweepSFminusTmcSix{}$ at mc\,=\,5--7,
peaking at $\StatSweepSFminusTExact{}$ at the exact/synonym default.
At low mc\,=\,0--4, any-length substring matches inflate all three models'
scores equally, compressing the absolute gap while leaving the ordering
unchanged.

\section{Metric and Style Feature Definitions}
\label{app:metrics}

\paragraph{Accuracy (M3).}
A response is correct if the content of the \texttt{<diagnosis>} tag matches the
ground-truth disease name via bidirectional substring match against a synonym
list: either the tag content contains the ground-truth name, or vice versa
(default; mc\,=\,None). Responses with no \texttt{<diagnosis>} tag score 0.

\paragraph{Mentioned (M1).}
The ground-truth disease name appears anywhere in the full response text
(case-insensitive). This credits passing mentions in rejected hypotheses and is
reported for completeness only.

\paragraph{Self-correction count.}
Number of self-correction phrases (e.g., ``however'', ``on the other hand'',
``wait'', ``actually'', ``let me reconsider'') per response, counted in the
\texttt{<think>} chain.

\paragraph{Explicit uncertainty count.}
Number of hedging phrases indicating explicit uncertainty (e.g., ``I'm not sure'',
``it's possible that'', ``could be'', ``uncertain'') per response.

\paragraph{Confident assertions.}
Number of assertive commitment phrases (e.g., ``the diagnosis is'', ``this is
clearly'', ``the patient has'', ``consistent with'') per response.

\paragraph{SC ratio.}
Ratio of mean self-correction count on incorrect responses to mean self-correction
count on correct responses. A ratio of 1.0 indicates equal hedging regardless of
correctness; higher values indicate elevated hedging on wrong answers.

%==============================================================================
\section{Reproducibility Details}
\label{app:reproducibility}
%==============================================================================
All code, trained LoRA adapters, evaluation result files, and filtered training
datasets are available \href{https://github.com/joetheguide2/hindsight/tree/main}{here}.

\begin{table*}[htbp]
\small\centering
\setlength{\tabcolsep}{4pt}
\renewcommand{\arraystretch}{1.15}
\begin{tabularx}{\textwidth}{@{}lX@{}}
\toprule
\multicolumn{2}{@{}l}{\textit{Dataset — ZebraMap~\cite{zebramap2026}}} \\
\quad Eval set  & \ConstZebraMapEvalDiseases{} diseases $\times$ \ConstZebraMapEvalCasesPerDisease{} cases $=$ \ConstZebraMapEvalCasesTotal{} \\
\quad Train set & \ConstZebraMapTrainDiseases{} diseases, \ConstZebraMapTrainCases{} cases (4 quality filters) \\
\quad Splits    & Raw CoT $\to$ Student; GT-redacted CoT $\to$ StudentF \\
\midrule
\multicolumn{2}{@{}l}{\textit{Hindsight trace generation}} \\
\quad Teacher   & \ConstTeacherBase{}~\cite{deepseek-r1-2025} (GGUF \ConstTeacherQuant{}) via llama.cpp \\
\quad Input     & Case narrative + ground-truth label \\
\quad Params    & max 4{,}096 tokens; temperature 0.6 \\
\midrule
\multicolumn{2}{@{}l}{\textit{Student training}} \\
\quad Base      & \ConstStudentBase{}-Instruct~\cite{qwen25-2024} (Unsloth, 4-bit \ConstStudentQuant{}) \\
\quad LoRA      & $r{=}\ConstLoRArank{}$, $\alpha{=}\ConstLoRAalpha{}$, all 7 projection layers \\
\quad Schedule  & 3 epochs; batch 32 (4 $\times$ 8 grad.\ accum.); lr $2{\times}10^{-4}$ cosine \\
\quad Other     & max 8{,}192 tokens; Paged AdamW 8-bit; wd $10^{-3}$; seed 3{,}407 \\
\midrule
\multicolumn{2}{@{}l}{\textit{Evaluation}} \\
\quad Prompt    & Zero-shot; case narrative only \\
\quad Teacher   & llama.cpp; max\_tokens 4{,}096; temp 0.6 \\
\quad Students  & Transformers/Unsloth; max\_new\_tokens 2{,}048; temp 0.6; top-$p$ 0.9 \\
\quad Base      & Full floating-point (conservative for fine-tuned models) \\
\quad BM25      & $k_1{=}1.5$, $b{=}0.75$, $k{=}1$; accuracy $=$ \StatBMTwentyFiveBest{} \\
\midrule
\multicolumn{2}{@{}l}{\textit{Statistical methodology}} \\
\quad Clusters  & \ConstClusterN{} diseases $\times$ \ConstZebraMapEvalCasesPerDisease{} cases each \\
\quad Perm.\ test & $B{=}\ConstPermB{}$; two-tailed; Holm--Bonferroni corrected \\
\quad Bootstrap & $B{=}\ConstBootstrapB{}$; 95\% percentile; effects: RD, Cohen's $h$, OR \\
\quad ICC(1)    & Shrout \& Fleiss (1979) Case-1; disease as grouping factor \\
\quad Freq.     & OLS on $\log(1{+}\text{freq})$ and Spearman $\rho$ \\
\quad Calibration & Grouped $k$-fold logistic regression; OOF AUC per model \\
\quad Stats file & Auto-generated to \texttt{generated\_stats.tex}; \texttt{\textbackslash input}ted at compile time \\
\bottomrule
\end{tabularx}
\caption{Reproducibility settings.}
\label{tab:repro}
\end{table*}

\end{document}